\documentclass[11pt, a4paper]{article}
\usepackage{latexsym}
\usepackage[fleqn]{amsmath}
\usepackage{graphicx}
\usepackage[fleqn]{amsmath}
\usepackage[varg]{txfonts}
\usepackage[english]{babel}
\usepackage[hyperindex=false,
    pageanchor=false,
    pdfmenubar=false,
    pdfpagelabels=false, breaklinks=true]{hyperref}
\usepackage{natbib}
\bibpunct{(}{)}{,}{a}{,}{,}

\usepackage{setspace}
\usepackage{epstopdf}
\usepackage{setspace}
\usepackage{graphicx}
\usepackage{threeparttable}
\newcommand{\argmin}{\operatornamewithlimits{argmin}}
\newcommand{\mb}{\mathbb} 
\newcommand{\vc}{\mathbf} 

\newcommand{\mc}{\mathcal}
\newcommand{\mr}{\mathrm}

\newtheorem{thm}{Theorem}[section]
\newtheorem{lem}{Lemma}[section]
\newtheorem{cor}{Corollary}[section]

\newtheorem{defn}{Definition}[section]
\newtheorem{Exm}{Example}[section]

\begin{document}
\title{Robust Kernel (Cross-) Covariance  Operators in Reproducing Kernel Hilbert Space toward Kernel Methods} 
\author{\textbf{  Md. Ashad Alam$^{1,2}$,  Kenji Fukumizu$^3$ and Yu-Ping Wang$^1$}  \\ $^{1}$Biomedical Engineering Department, Tulane University\\
 New Orlenas, LA 70118,USA\\
$^2$Department of Statistics, Hajee Mohammad Danesh Science and Technology\\ University Dinajpur 5200, Bangladesh\\
$^3$The Institute of Statistical Mathematics, Tachikawa, Tokyo 190-8562, Japan}       
\date{}
\maketitle
\begin{abstract}
To the best of our knowledge, there are no general well-founded robust methods for statistical unsupervised learning.  Most of the unsupervised methods  explicitly or implicitly  depend  on  the kernel covariance operator (kernel CO) or  kernel cross-covariance operator (kernel CCO). They are sensitive to contaminated data, even when using bounded positive definite kernels. First, we propose   robust kernel covariance operator (robust kernel CO) and robust  kernel cross-covariance operator (robust kernel CCO)  based on a generalized loss function instead of the quadratic loss function. Second, we propose influence function  of classical kernel canonical  correlation analysis ({\em classical kernel CCA}). Third, using this influence function, we propose a visualization method  to detect influential observations from two sets of data. Finally, we propose a  method based on robust kernel CO and robust kernel CCO, called {\em robust kernel CCA}, which is designed for contaminated data and  less sensitive to noise than {\em classical kernel CCA}. The principles we describe also apply  to  many kernel methods which must deal with the issue of kernel CO or kernel CCO. Experiments on synthesized and imaging genetics analysis demonstrate that the proposed visualization and  {\em robust  kernel CCA} can be applied effectively to   both ideal data and contaminated data. The robust methods show the superior performance over the  state-of-the-art methods. 
\end{abstract}  

\section{Introduction}
\label{sec:Intro}
\label{sec:Intro}
The incorporation of  various unsupervised learning methods for multiple data sources into genomic analysis is a rather  recent topic. Using the dual representations, the task of learning with multiple data sources is related to the kernel-based data fusion, which has been actively studied in the last decade \cite{Back-08,Steinwart-08,Hofmann-08}. Kernel fusion in unsupervised learning has a close connection with unsupervised kernel methods. As unsupervised kernel methods, kernel principal component analysis \citep[kernel PCA]{Schlkof-kpca,Ashad-14}, kernel canonical correlation analysis \citep[classical kernel CCA]{Akaho,Back-02}, weighted multiple kernel CCA and others have been extensively studied in unsupervised kernel fusion for decades \citep{Yu-11}. But these methods  are not robust; these are sensitive to contaminated data. Even though a number  of researches has been done on robustness issue for supervised learning, especially  support vector  machine for classification and regression \citep{Christmann-04,Christmann-07,Debruyne-08}, there are no general well-founded robust methods for unsupervised learning.

Robustness is an essential and challenging  issue in statistical machine learning for  multiple sources data analysis. Because {\em outliers}, data  that cause surprise in relation to the majority of the data, are  often occur in the real data. Outliers may be right, but we need to  examine for transcription errors. They can play havoc with classical statistical methods or statistical machine learning methods. To overcome this problem, since 1960 many robust methods have been developed, which are  less sensitive to outliers.  The  goals of robust statistics are  to use the methods  from  the bulk of the data and indicate the points deviating from the original pattern  for further investment \citep{Huber-09,Hampel-11}. In recent years, a robust kernel density estimation (robust kernel DE) has been proposed \cite{Kim-12}, which is less sensitive than the kernel density estimation.  To the  best of our knowledge,  two spacial robust  kernel PCA methods have been  proposed based on  weighted eigenvalues decomposition  \citep{Huang-KPCA} and  spherical kernel PCA \citep{Debruyne-10}. They show that the influence function (IF), a well-known measure of robustness, of kernel PCA can be  arbitrary large for unbounded kernels.

During   the last ten years,  a number of  papers have been about  the properties of kernel  CCA, CCA using positive definite kernels, called {\em classical kernel CCA} and its variants have been proposed \citep{Fukumizu-SCKCCA, Hardoon2009, Otopal-12,Ashad-15}.  Due to the properties of eigen decomposition it is still a well applied methods for multiple souses data analysis. In recent years, two  canonical correlation analysis (CCA) methods based on Hilbert-Schmidt independence criterion (hsicCCA) and  centered kernel target alignment (ktaCCA) have been proposed by \cite{Chang-13}. These methods are able to extract nonlinear structure of the data as well. Due to the gradient based optimization, these methods are not able to extract all canonical variates using the same initial value and do not work for high dimensional datasets.  For more details, see  Section \ref{sec:classification}.  An empirical comparison and   sensitivity analysis   for robust linear CCA and classical kernel CCA is also discussed,  and gives  similar  interpretation as kernel PCA for  kernel CCA without any theoretical results \citep{Ashad-10}.

Most of the  kernel methods explicitly or implicitly depend on kernel covariance operator (kernel CO) or kernel cross-covariance operator (kernel CCO). Among others, these are  most useful tools of unsupervised kernel methods but have not been robust yet. They can be formulated as an empirical optimization problem  to achieve robustness by combining empirical optimization problem with ideas of Huber or Hampel's M-estimation model \citep{Huber-09,Hampel-11}. The robust kernel CO and  robust kernel CCO can be computed efficiently via a kernelized iteratively re-weighted least square (KIRWLS) problem.  In robust kernel DE based on robust kernel mean elements (robust kernel ME) is used KIRWLS in reproducing kernel Hilbert space (RKHS) \citep{Kim-12}. \cite{Debruyne-10}  have proposed  a visualization methods for detecting influential observations from one set of the data using IF of kernel PCA.  In addition, \cite{Romanazii-92} has  proposed the IF  of canonical correlation and canonical vectors of linear CCA but the IF of classical kernel CCA and  any robust kernel CCA have not been proposed, yet.  All of these considerations motivate us to conduct  studies on robust kernel CCO toward kernel unsupervised methods. 

Contribution of this paper is fourfold. First, we propose   robust kernel  CO and robust kernel CCO   based on generalized loss function instead of the quadratic loss function. Second, we propose IF of  classical kernel CCA: kernel canonical correlation (kernel CC) and kernel canonical variates (kernel CV). Third, to detect influential observations from multiple sets of data, we propose a visualization method using the inflection function of kernel CCA. Finally, we propose a  method based on robust kernel CO and robust kernel CCO, called {\em robust kernel CCA}, which is less sensitive than classical kernel CCA. Experiments on synthesized and imaging genetics analysis demonstrate that the proposed visualization and  robust {\em kernel CCA} can be applied effectively to both ideal data (ID) and contaminated data (CD).

The remainder of this paper is organized as follows. In the next Section, we provide a brief review of  kernel ME,  kernel  CCO, robust kernel ME, robust kernel CO, robust kernel CCO and robust Gram matrices with algorithms. In Section $3$,  we discuss in  brief  the IF, IF of kernel ME and IF of kernel CO and  kernel CCO. After a brief review of classical  kernel CCA in  Section \ref{sec:CKCCA}, we propose the  IF of classical  kernel CCA: kernel CC and  kernel CV in Section \ref{sec:IFKCCA}. The   {\em robust kernel CCA} is proposed in Section \ref{sec:RKCCA}. In Section $5$, we describe  experiments conducted on both synthesized data and  the imaging genetics analysis with a visualizing method. In  Appendix, we discuss the results in detail.

\section{Classical and robust kernel  (cross-) covariance operator in RKHS}
Kernel ME, kernel CO and kernel CCO  with positive definite kernel have been extensively applied to nonparametric statistical inference  through representing distribution  in the form of means and covariance in RKHS \citep{Gretton-08, Fukumizu-08,Song-08, Kim-12,Gretton-12}. Basic notion of   kernel MEs, kernel CO and kernel CCO with its robustness through IF are briefly discussed below.
\subsection{Classical kernel (cross-) covariance operator}
\label{sec:kernel CCO}
Let $F_X$, $F_Y$  and $F_{XY}$ be  the probability measure on $\mc{X}$, $\mc{Y}$ and  $ \mc{X}\times\mc{Y}$, respectively.
Also let $X_1, X_2, \ldots, X_n$,;  $ Y_1, Y_2, \ldots, Y_n$ and $(X_1, Y_1),(X_2, Y_2), \ldots, (X_2, Y_2)$ be the random  sample from the distribution  $F_X$, $F_Y$  and $F_{XY}$, respectively. A symmetric kernel $k(\cdot,\cdot)$ defined on a space  is called {\em positive definite kernel} if   the Gram matrix $(k(X_i, X_j))_{ij}$ is positive semi-definite \citep{Aron-RKHS}.  By the reproduction properties and kernel trick, the kernel can evaluate the inner product of any two feature vectors efficiently without knowing an explicit form of either the {\em feature map} ($\Phi(\cdot) = k(\cdot, X), \forall X\in \mc{X}$) or {\em feature spaces} ($\mc{H}$). In addition,  the computational cost does not depend on  dimension of the original space after computing the  Gram matrices \citep{Fukumizu-14,Ashad-14}.

   A mapping $\mc{M}_X:= \mb{E}_X[\Phi(X)] =  \mb{E}_X[k(\cdot, X)]$ with $\mb{E}_X[k(X, X)] < \infty$ is an  element of the RKHS $\mc{H}_X$. By the reproducing property  with $X\in \mc{X}$, {\em kernel mean elements} is defined as
\[\langle \mc{M}_X, f \rangle_{\mc{H}_X} = \langle \mb{E}_X[k(\cdot, X)], f \rangle = \mb{E}_X\langle  k(\cdot, X), f \rangle_{\mc{H}_X}=\mb{E}_X[f(X)],$
for all    $f\in \mc{H}_X\]. 
Given  an  independent and identically distributed sample, the mapping $m_X=\frac{1}{n}\sum_{i=1}^n\Phi(X_i)=  \frac{1}{n} \sum_{i=1}^n k(\cdot, X_i)$ is an  empirical element of the RKHS, $\mc{H}_X$,
$\langle m_X, f \rangle_{\mc{H}_X}= \langle  \frac{1}{n}\sum_{i=1}^n k(\cdot, X_i), f\rangle =  \frac{1}{n}\sum_{i=1}^n  f(X_i).
$
The  sample kernel ME of  the feature vectors $\vc{\Phi}(X_i)$  can be regraded as  a solution to the empirical risk optimization problem \citep{Kim-12}
\begin{eqnarray}
\label{EROP1}
\argmin_{f\in \mc{H}_X} \frac{1}{n}\sum_{i=1}^n\| \vc{\Phi}(X_i)- f\|^2_{\mc{H}_X}.
\end{eqnarray} 
Similarly, we can define kernel CCO  as  an empirical risk optimization problem.  An operator, $\Sigma_{YX} := \mc{H}_X \to \mc{Y}_Y$ with $\mb{E}_X[k_X(X, X)] < \infty$, and $\mb{E}_Y[k_Y(Y, Y)] < \infty$, by the reproducing property  which is defined  as
\begin{eqnarray}
\langle f_X,  \Sigma_{YX}f_X\rangle_{\mc{H}_Y} &=& \mb{E}_{XY}\left[\langle f_X, k_X(\cdot, X) - \mc{M}_X \rangle_{\mc{H}_X} \langle f_Y, k_Y(\cdot, Y) - \mc{M}_Y \rangle_{\mc{H}_Y} \right]\nonumber\\&=& \mb{E}_{XY}\left[(f_X(X) - E_X[f(X)]) (f_Y(Y) - E_Y[f(Y)])\right] \nonumber
\end{eqnarray} 
and called {\em kernel CCO}. Given the  pair of  independent and identically distributed sample, $(X_i, Y_i)_{i=1}^n$,  the kernel CCO is an operator of the RKHS, $\mc{H}_X\otimes\mc{H}_Y$,
 Eq. (\ref{EROP1}) becomes 
\begin{eqnarray}
\label{EROP2}
\argmin_{\Sigma_{XY}\in \mc{H}_X\otimes \mc{H}_Y}\frac{1}{n} \sum_{i=1}^n \| \vc{\Phi}_c(X_i) \otimes \vc{\Phi}_c(Y_i) - \Sigma_{XY}\|^2,
\end{eqnarray} 
where $\vc{\Phi}_c(X_i)= \vc{\Phi}(X_i)-\frac{1}{n}\sum_{b=1}^n \vc{\Phi}(X_b)$. and the kernel covariance operator at point $(X_i, Y_i)$ is then 
\[\hat{\Sigma}_{YX} (X_i, Y_i)= (k_X(X_i, X_b)- \frac{1}{n}\sum_{b=1}^nk_X(\cdot, X_b))\otimes (k_Y(Y_i, Y_d)- \frac{1}{n}\sum_{d=1}^n k_Y(Y_i, Y_d)).\] Special case, if Y is equal to X,  gives kernel CO.

\subsection{Robust kernel (cross-) covariance operator}
It  is known that (as in  Section \ref{sec:kernel CCO})  the   kernel ME is the   solution to the empirical risk optimization problem, which are the least square  type  estimators.  This type of   estimators are sensitive  to the presence of {\em outliers} in the features, $\Phi(X_i)$. In recent years, the robust kernel ME has been proposed for density estimation  \citep{Kim-12}. Our goal is to extend this notion to kernel CO and kernel CCO. To do these, we estimates kernel CO and kernel CCO   based on robust loss functions, {\em M-estimator}, and called, {\em robust kernel CO} and  {\em robust kernel CCO}, respectively.  Most common example of robust loss functions, $\zeta(t)$ on $t \geq 0$, are  Huber's or Hampel's loss function.  Unlike the quadratic loss function, the derivative of these loss functions  is bounded \citep{Huber-09, Hampel-86}. The  Huber's   function  is  defined as
\begin{eqnarray}
\zeta(t)=
\begin{cases}
	 t^2/2\qquad  \qquad ,0\leq t\leq c  
\\
 ct-c^2/2\qquad  ,c\leq t \nonumber 
\end{cases}
\end{eqnarray}
and Hampel's  function is defined as
\begin{eqnarray}
\zeta(t)=
\begin{cases}
	 t^2/2\qquad  \qquad\qquad\qquad ,0\leq t\le c_1  
\\
 c_1t-c_1^2/2\qquad\qquad  ,c_1\leq t < c_2
\\
c_1(t-c_3)^2/2 (c_2-c_3)+ c_1(c_2+c_3-c_1)/2\qquad  ,c_2\leq t < c_3\\
 c_1(c_2+c_3-c_1)/2 \qquad\qquad  ,c_3\leq t. \nonumber
\end{cases}
\end{eqnarray}

The basic assumptions are: (i) $\zeta$ is non-decreasing, $\zeta(0)=0$ and  $\zeta(t)/t \to 0$ as $t\to 0$ (ii) $\varphi(t)=\frac{\zeta^\prime(t)}{t}$  exists and is finite, where $\zeta^\prime(t)$ is derivative of $\zeta(t)$. (iii) $\zeta^\prime(t)$ and $\varphi(t)$ are continuous and bounded (iv) $\varphi(t)$ is Lipschitz  continuous. Huber's loss function  as well as others hold  for all of these assumptions \citep{Kim-12}.

Given weights of  robust kernel ME, $\vc{w}=[ w_1, w_2, \cdots, w_n]^T$, of  a set of observations,   the points $\vc{\Phi}_c(X_i):= \vc{\Phi}(X_i) - \sum_{a=1}^nw_a \vc{\Phi}(X_a)$  are centered and the centered Gram matrix is $\tilde{K}_{ij}=(\vc{H}\vc{K}\vc{H}^T)_{ij}$,  
where $\vc{1}_n=[1, 1, \cdots, 1]^T$ and $\vc{H}=\vc{I}- \vc{1}_n\vc{w}^T$.

 Eq. (\ref{EROP2}) can be written as    
\begin{eqnarray}
\label{REROP1}
\argmin_{f\in \mc{H}_X} \frac{1}{n}\sum_{i=1}^n \zeta(\| \vc{\Phi}_c(X_i) \otimes \vc{\Phi}_c(Y_i) - \Sigma_{XY}\|).
\end{eqnarray}
As in \citep{Kim-12}, Eq. (\ref{REROP1})  does not has a closed form solution, but   using the kernel trick the classical re-weighted least squares (IRWLS) can be extended to a RKHS. The solution is then,
\[\widehat{\Sigma}_{XY}^{(h)}= \sum_{i=1}^n w_i^{(h-1)}\tilde{k}(X, X_i)\tilde{k}(Y, Y_i),\]
where $w_i^{(h)}=\frac{\varphi(\|\vc{\Phi}_c(X_i)\otimes \vc{\Phi}_c(Y_i) - \Sigma_{XY}\|_{\mc{H}_X\otimes\mc{H}_Y}}{\sum_{b=1}^n\varphi(\| \vc{\Phi}_c(X_b)\otimes\vc{\Phi}_c(Y_b)- \Sigma_{XY}\|_{\mc{H}_X\otimes\mc{H}_Y})}\,, \rm{and} \, \varphi(x)=\frac{\zeta^\prime(x)}{x}.$

The algorithms of estimating robust Gram matrix and robust kernel CCO  are given in Figure \ref{Robust.K.M} and  in Figure \ref{Robust.cross.covariance.M}, respectively.
\begin{figure}
\hrule
Input: $D=\{\vc{X}_1, \vc{X}_2 ,\ldots \vc{X}_n \}$ in $\mb{R}^m$. The kernel  matrix $\vc{K}$ with kernel $k$ and $\vc{K}_{\vc{X}_i}= k(\cdot, \vc{X}_i)$.   Threshold $TH$, (e.g., $10^{-8}$). The objective function of robust mean element is
\[ M_R=\arg \min_{f\in \mc{H}} J(f),\qquad \rm {where}  \,
J(f)= \frac{1}{n}\sum_{i=1}^n \rho(\|K_{X_i} - f\|_{\mc{H}} )\]
\begin{enumerate}
\item[] Do the following steps until:
\[ \frac{|J(M_R^{(h+1)})- J(M_R^{(h)})|}{J(M_R^{(h)})} < TH,\]
where$ M_R^{(h)}=\sum_{i=1}^nw_i\vc{K}_{\vc{X}_i}, \,
w_i^{(h)}=\frac{\varphi(\|\vc{K}_{\vc{X}_i}- M_R^{(h)}\|_\mc{H})}{\sum_{i=1}^n\varphi(\|\vc{K}_{\vc{X}_i}- M_R^{(h)}\|_\mc{H})}\,, \rm{and} \, \varphi(x)=\frac{\xi^\prime(x)}{x}$
\begin{itemize}
\item[(1)] Set $h=1$ and $w_i^{(0)}=\frac{1}{n}$.
\item[(2)] Solve  $w_i^{(h)}=\frac{ \varphi(\epsilon_i^{[h]})}{\sum_i^n\varphi(\epsilon_i^{[h]})}$ and make a vector $\vc{w}$ for  $i=1, 2, \cdots n$.
\item[(3)] Update the mean element,  $M_R^{(h+1)}=  [\vc{w}^{(h)}]^T\vc{K}$.
\item[(4)] Update error,  $\epsilon^{[h+1]} =(\rm{diag}(\vc{K})- 2[\vc{w}^{(h)}]^T \vc{K}+ [\vc{w}^{(h)}]^T \vc{K}[\vc{w}^{(h)}]^T\vc{1}_n )^{\frac{1}{2}} $.
\item[(5)]  Update $h$ as $h+1$.
\end{itemize}
\end{enumerate}
Output:  the centered robust kernel matrix, $\tilde{\vc{K}}_R= \vc{H}\vc{K}\vc{H}^T$ where $\vc{H}=\vc{I}_n- \vc{1}_n \vc{w}^T$
\vspace*{3mm}
\hrule
\caption{ The algorithm of estimating centered kernel matrix using robust kernel mean element.}
\label{Robust.K.M}
\end{figure}

\begin{figure}
\hrule
Input: $D=\{(\vc{X}_1,\vc{Y}_1),  (\vc{X}_2, \vc{Y}_2), \ldots (\vc{X}_n, \vc{Y}_n) \}$. The robust centered kernel  matrix $\tilde{\vc{K}}_X$ and $\tilde{\vc{K}}_Y$  with kernel $k_X$ and $k_Y$, $\tilde{K}_{Xi}$ and,   $\tilde{K}_{Yi}$   are the $i$th column of the $\tilde{\vc{K}}_X$ and $\tilde{\vc{K}}_Y$, respectively. Also  define $\tilde{\vc{K}}_{\vc{X}_i}= k_X (\cdot, \vc{X}_i)$ and  $\tilde{\vc{K}}_{\vc{Y}_i}= k_y(\cdot, \vc{Y}_i)$. Threshold $TH$ (e.g., $10^{-8}$). The objective function of robust  cross-covariance operator  is
\[ \hat{\Sigma}_R=\arg \min_{\vc{A}\in {\mc{H}_X\otimes\mc{H}_Y}}  J(\vc{A}),\qquad \rm {where}  \,
J(\vc{A})= \frac{1}{n}\sum_{i=1}^n \rho(\|\vc{B}_i - \vc{A}\|_{\mc{H}_X\otimes\mc{H}_Y}),\]
\[\vc{B}_i=\tilde{\Phi}(\vc{X}_i)\otimes\tilde{\Phi}(\vc{Y}_i)^T=\tilde{\vc{K}}_{\vc{X}_i}\otimes \tilde{\vc{K}}_{\vc{Y}_i}\]
\begin{enumerate}
\item[] Do the following steps until:
\[ \frac{|J\Sigma_R^{(h+1)})- J(\Sigma_R^{(h)})|}{J(\Sigma_R^{(h)})} < TH,\]
where\, $ \hat{\Sigma}_R^{(h)}=\sum_{i=1}^nw_i^{(h-1)} \vc{B}_i,\,  
 w_i^{(h)}=\frac{\varphi(\|\vc{B}_i-  \hat{\Sigma}_R^{(h)}\|_{ \mc{H}_X\otimes\mc{H}_Y})}{\sum_{i=1}^n  \varphi(\|\vc{B}_i- \hat{\Sigma}_R^{(h)}\|_{\mc{H}_X\otimes\mc{H}_Y})}\, \rm{ and} \, \varphi(x)=\frac{\xi^\prime(x)}{x} $
\begin{itemize}
\item[(1)] Set $h=1$, and $ w_i^{(0)}=\frac{1}{n}$
\item[(2)] Solve  $w_i^{(h)}=\frac{ \varphi(\epsilon_i^{[h]})}{\sum_i^n\varphi(\epsilon_i^{[h]})}$ and make a vector $\vc{w}$ for  $i=1, 2, \cdots n$.
\item[(3)]  Calculate a $n^2\times 1$ vector, 
$\vc{v}^{(h)}= \vc{B} \vc{w}^{(h)}$ and make a $n\times n$ matrix $\vc{V}^{(h)}$, where $\vc{B}$ is $n^2\times n$ matrix  that $i$th column consists of all elements of the $n\times n$ matrix  $\vc{B}_i$.
\item[(4)]  Update the robust covariance,  $\hat{\Sigma}_R^{(h+1)}= \sum_i^nw_i^{(h)}\vc{B}_i= \vc{V}^{(h)}$.
\item[(5)] Update error,  $\epsilon^{[h+1]} =(\rm{diag}(\tilde{\vc{K}}_X\tilde{\vc{K}}_Y)- 2[\vc{w}^{(h)}]^T \tilde{\vc{K}}_X \tilde{\vc{K}}_X+ [\vc{w}^{(h)}]^T \tilde{\vc{K}}_X\tilde{\vc{K}}_Y[\vc{w}^{(h)}]^T\vc{1}_n )^{\frac{1}{2}} $.
\item[(6)]  Update $h$ as $h+1$.
\end{itemize}
\end{enumerate}
Output:  the robust cross-covariance operator.
\hrule
\caption{ The algorithm of estimating robust cross-covariance operator.}
\label{Robust.cross.covariance.M}
\end{figure}

\section{Influence function of kernel (cross-) covariance operator}
To define the notion of robustness in statistics, different  approaches have been proposed  science 70's decay for examples, the {\em minimax approach} \citep{Huber-64}, the {\em sensitivity curve} \citep{Tukey-77}, the {\em influence functions} \citep{Hampel-74,Hampel-86} and in  the finite sample {em breakdown point} \citep{Dono-83}. Due to simplicity, IF  is the most useful approach in  statistics and in statistical supervised learning \cite{Christmann-07,Christmann-04}. In this section, we briefly discuss the notion of IF, IF of kernel ME, IF of kernel CO and kernel CCO. (For  details  see in  Appendix).

Let ($\Omega$, $\mc{A}$, $\mb{P}$) be a probability space and $(\mc{X}, \mc{B})$  a measure space. We want to estimate the parameter $\theta \in \Theta$  of a distribution  $F$ in $\mc{A}$.  We assume that exists a functional $R: \mc{D}(R) \to \mb{R}$, where $ \mc{D}(R)$  is the set of all  probability distribution in $\mc{A}$. Let $G$ be some distribution in $\mc{A}$. If data do not fallow the model $F$ exactly but  slightly  going toward $G$,  the  G\^{a}teaux Derivative at $F$ is given by 
\begin{eqnarray}
\lim_{\epsilon\to 0 }\frac{R[(1-\epsilon)F+\epsilon G] - R(F)}{\epsilon}
\end{eqnarray}
Suppose $x\in \mc{X}$ and $G=\Delta_x$ is the probability measure which gives mass $1$  to $\{x\}$. The {\em influence function} (special case of G\^{a}teaux Derivative) of $R$ at $F$ is defined by
\begin{eqnarray}
IF(x, R, F)=\lim_{\epsilon\to 0 }\frac{R[(1-\epsilon)F+\epsilon\Delta_x] - R(F)}{\epsilon}
\end{eqnarray}
provided that the limit exists. It can be intuitively interpreted as a suitably normalized asymptotic influence of outliers on the value of an estimate or test statistic.

There are three properties of IF: gross error sensitivity, local shift sensitivity and  rejection point.  They  measured the worst effect of gross error, the worst effect of rounding error and  rejection point. For a scalar, we just define influence function (IF) at a fixed point. But if the estimate is a function, we are able to express the change of the function value at every points \citep{Kim-12}.

\subsection{Influence function  of kernel mean element and kernel cross-raw moment}
For a scalar we just define IF  at a fixed point. But if the estimate is a function,  we are able to express the change of the function value at every point.

Let the  cross-raw moments  
\[R(F_{XY}) = \mb{E}_{XY}[ \langle k_X(\cdot, X), f \rangle_{\mc{H}_X} \langle k_Y(\cdot,Y), g \rangle_{\mc{H}_Y}]= \mb{E}_{XY}[f(X)g(Y)]= \int f(X) g(Y)dF_{XY}.\]

The  IF of $R(F_X)$ at $Z^\prime=(X^\prime,Y^\prime) $ for every points $(\cdot)$ is given by

\[IF(\cdot, Z^\prime, R, F_{X Y}) = k_X(\cdot, X^\prime) k_Y(\cdot, Y^\prime) -\mb{E}_{XY}[ \langle k_X(\cdot, X), f \rangle_{\mc{H}_X} \langle k_Y(\cdot,Y), g \rangle_{\mc{H}_Y}],\]
$ \forall k_X(\cdot, X) \in  \mc{H}_X,\,  k_Y(\cdot, Y) \in \mc{H}_Y$, which is estimated  with the pairs of data points $ (X_1 Y_1), (X_2,Y_2), \cdots, (X_n,Y_n) \in  \mc{X}\times  \mc{Y}$ at any evaluated point $ (X_i,Y_i) \in \mc{X}\times  \mc{Y} $
\begin{eqnarray}
k_X(X_i, X^\prime) k(Y_i, Y^\prime)- \frac{1}{n}\sum_{a=1}^n k_X(X_i, X_a)k_Y(Y_i,Y_a)  \qquad \forall \, k_X(\cdot, X_a) \in \mc{H}_\mc{X},\,   k_Y(\cdot, Y_a) \in \mc{H}_\mc{Y}. \nonumber
\end{eqnarray} 

\subsection{Influence function of  complicated statistics}
The IF of  complicated statistics,  which are functions of simple statistics,  can be calculated with the chain rule. Say 
$R(F)=a\{R_1(F), ....., R_s(F)\}$, then
 \begin{eqnarray}
IF_R(z)=\sum_{i=1}^s \frac{\partial a}{\partial R_i}IF_{R_i }(z). \nonumber
 \end{eqnarray}
It can also be used to find the IF for a transformed statistic, given the influence function  for the statistic itself.

The IF of kernel CCO, $R(F_{XY})$, with joint distribution, $F_{XY}$, using complicated statistics at  $Z^\prime=(X^\prime,Y^\prime)$ is given by
\begin{align*}
&\rm{IF}(\cdot, Z^\prime, R, F_{XY}) \nonumber \\
&= \langle k_X(\cdot, X^\prime)-\mc{M}[F_X], f \rangle_{\mc{H}_X} \langle k_Y(\cdot,Y^\prime)\mc{M}[F_Y], g \rangle_{\mc{H}_Y} 
\\ &-\mb{E}_{XY}[ \langle k_X(\cdot, X)-\mc{M}[F_X], f \rangle_{\mc{H}_X} \langle k_Y(\cdot,Y)-\mc{M}[F_Y], g \rangle_{\mc{H}_Y}],
\end{align*}
which is estimated with the data points $(X_1 Y_1), (X_2, Y_2), \cdots, (X_n, Y_n) \in \mc{X}\times \mc{Y}$ for every $Z_i = (X_i, Y_i)$ as
\begin{multline}
\widehat{\rm{IF}}( Z_i, Z^\prime, R, F_{XY})  \\
=  [k_X(X_i, X^\prime)-\frac{1}{n}\sum_{b=1}^n k_X(X_i, X_b)] [k_Y(Y_i, Y^\prime)\\-\frac{1}{n}\sum_{b=1}^n k_Y(Y_i, Y_b)] - 
\frac{1}{n}\sum_{d=1}^n[k_X(X_i, X_d)-\frac{1}{n}\sum_{b=1}^n k_X(X_i, X_b)] [k_Y(Y_i, Y_d)-\frac{1}{n}\sum_{b=1}^n k_Y(Y_i, Y_b)]. \nonumber
\end{multline}
For the  bounded kernels, the above  IFs have  three properties: gross error sensitivity, local shift sensitivity  and rejection point. It is not true for the unbounded kernels, for example, liner and polynomial kernels. We can make similar conclusion for the kernel covariance operator.

\section{Classical and robust kernel canonical  correlation analysis}
In this Section, we  review classical kernel CCA and propose the IF  and empirical IF (EIF)  of kernel CCA. After that we  propose a {\em robust kernel CCA} method based  on robust kernel CO and robust kernel CCO. 

\subsection{Classical kernel CCA}
\label{sec:CKCCA}
{\em Classical kernel CCA} has been proposed as a nonlinear extension of linear CCA \citep{Akaho,Lai-00}. \cite{Back-02} has extended the  {\em classical kernel CCA}  with efficient computational algorithm, incomplete Cholesky factorization.  Over the last decade, {\em classical kernel CCA} has been used for various purposes including preprocessing for classification, contrast function of independent component analysis, test of independence between two sets of variables, which has been applied in many domains such as genomics, computer graphics and computer-aided drug discovery and  computational biology \citep{Alzate2008,Hardoon2004,Huang-2009}.  Theoretical results on the convergence of kernel CCA have also been  obtained \citep{Fukumizu-SCKCCA,Hardoon2009}.

The aim of {\em classical kernel CCA}  is to seek  the sets of  functions in the RKHS for which the correlation (Corr) of  random variables  is maximized. The simplest case, given  two sets of random variables $X$  and $Y$ with  two   functions in the RKHS, $f_{X}(\cdot)\in \mc{H}_X$  and  $f_{Y}(\cdot)\in \mc{H}_Y$, the optimization problem of  the random variables $f_X(X)$ and $f_Y(Y)$ is
\begin{eqnarray}
\label{ckcca1}
\max_{\substack{f_{X}\in \mc{H}_X,f_{Y}\in \mc{H}_Y \\ f_{X}\ne 0,\,f_{Y}\ne 0}}\mr{Corr}(f_X(X),f_Y(Y)).
\end{eqnarray}
The optimizing functions $f_{X}(\cdot)$ and $f_{Y}(\cdotp)$ are determined  up to scale.

Using a  finite sample, we are able to estimate the desired functions. Given an i.i.d sample, $(X_i,Y_i)_{i=1}^n$ from a  joint distribution $F_{XY}$, by taking the inner products with elements or ``parameters" in the RKHS, we have  features
$f_X(\cdot)=\langle f_X, \Phi_X(X)\rangle_{\mc{H}_X}= \sum_{i=1}^na_X^ik_X(\cdot,X_i) $ and
 $f_Y(\cdot)=\langle f_Y, \phi_Y(Y)\rangle_{\mc{H}_Y}=\sum_{i=1}^na_Y^ik_Y(\cdot,Y_i)$, where $k_X(\cdot, X)$ and $k_Y(\cdot, Y)$ are the associated kernel functions for $\mc{H}_X$ and $\mc{H}_Y$, respectively. The kernel Gram matrices are defined as   $\vc{K}_X:=(k_X(X_i,X_j))_{i,j=1}^n $ and $\vc{K}_Y:=(k_Y(Y_i,Y_j))_{i,j=1}^n $.  We need the centered kernel Gram matrices $\vc{M}_X=\vc{C}\vc{K}_X\vc{C}$ and $\vc{M}_Y=\vc{C}\vc{K}_Y\vc{C}$, where $ \vc{C} = \vc{I}_n -\frac{1}{n}\vc{B}_n$ with  $\vc{B}_n = \vc{1}_n\vc{1}^T_n$ and $\vc{1}_n$ is the vector with $n$ ones. The empirical estimate of Eq. (\ref{ckcca1}) is then given by
\begin{eqnarray}
\label{ckcca6}
\max_{\substack{f_{X}\in \mc{H}_X,f_{Y}\in \mc{H}_Y \\ f_{X}\ne 0,\,f_{Y}\ne 0}}\frac{\widehat{\rm{Cov}}(f_X(X),f_Y(Y))}{[\widehat{\rm{Var}}(f_X(X))+\kappa\|f_X\|_{\mc{H}_X}]^{1/2}[\widehat{\rm{Var}}(f_Y(Y))+\kappa\|f_Y\|_{\mc{H}_Y}]^{1/2}} \nonumber 
\end{eqnarray}
where
\begin{align*}
& \widehat{\rm{Cov}}(f_X(X),f_Y(Y))
= \frac{1}{n} \vc{a}_X^T\vc{M}_X\vc{M}_Y \vc{a}_Y= \vc{a}_X^T\vc{M}_X\vc{W}\vc{M}_Y \vc{a}_Y , \\
& \widehat{\rm{Var}}( f_X(X))
=\frac{1}{n} \vc{a}_X^T\vc{M}_X^2 \vc{a}_X= \vc{a}_X^T\vc{M}_X \vc{W} \vc{M}_X \vc{a}_X, \,  \\ &\widehat{\rm{Var}}( f_Y(Y))=\frac{1}{n} \vc{a}_Y^T\vc{M}_Y^2 \vc{a}_Y= \vc{a}_Y^T\vc{M}_Y \vc{W} \vc{M}_Y\vc{a}_Y, 
\end{align*}
and  $\vc{W}$ is a diagonal matrix with elements  $\frac{1}{n}$, and  $\vc{a}_{X}$ and $\vc{a}_{Y}$ are the  directions of   $X$ and $Y$, respectively. The regularized coefficient $\kappa > 0$.

\subsubsection{Influence function of classical kernel CCA}
\label{sec:IFKCCA}
By using  the   IF results  of  kernel PCA, linear PCA and of   linear CCA,   we  can derive  the IF of kernel CCA: kernel CC and kernel CVs,
\begin{thm}
\label{TIFKCCA}
 Given two sets of random variables $(X, Y)$ having distribution   $F_{XY}$, the  influence function of kernel canonical correlation and canonical variate  at $Z^\prime = (X^\prime, Y^\prime)$ are given by 
\begin{multline}
\rm{IF} (Z^\prime, \rho_j^2)= - \rho_j^2 \bar{f}_{jX}^2(X^\prime) + 2 \rho_j \bar{f}_{jX}(X^\prime) \bar{f}_{jY}(Y^\prime)  - \rho_j^2 \bar{f}_{jY}^2(Y^\prime), \nonumber\\
\rm{IF} (\cdot, Z^\prime, f_{jX}) = -\rho_j (\bar{f}_{jY}(Y^\prime) - \rho_j \bar{f}_{jX}(X^\prime))\mb{L} \tilde{k} (\cdot, X^\prime) -  (\bar{f}_{jX}(X^\prime) \nonumber\\\qquad \qquad\qquad\qquad\qquad \qquad\qquad\qquad - \rho_j \bar{f}_{jY}(Y^\prime))\mb{L} \Sigma_{XY}\Sigma^{-1}_{YY} \tilde{k}_Y(\cdot,  Y^\prime)+\frac{1}{2}[1- \bar{f}^2_{jX}(X^\prime)]f_{jX}, \nonumber\\
 \rm{IF} (\cdot, Z^\prime, f_{jY})
= -\rho_j (\bar{f}_{jX}(X^\prime) - \rho_j \bar{f}_{jY}(Y^\prime))\mb{L} \tilde{k} (\cdot, Y^\prime)-  (\bar{f}_{jY}(Y^\prime)\\ - \rho_j \bar{f}_{jX}(X^\prime))\mb{L} \Sigma_{YX}\Sigma^{-1}_{XX} \tilde{k}_Y(\cdot, Y^\prime) +\frac{1}{2}[1- \bar{f}^2_{jY}(Y^\prime)]f_{jY}, \nonumber
\end{multline}
\rm{where} $\mb{L}= \Sigma_{XX}^{- \frac{1}{2}}(\Sigma_{XX}^{- \frac{1}{2}} \Sigma_{XY} \Sigma_{YY}^{-1} \Sigma_{YX}\Sigma_{XX}^{- \frac{1}{2}}-\rho^2\vc{I})^{-1}\Sigma_{XX}^{- \frac{1}{2}}$. 
\end{thm}
 To prove  Theorem \ref{TIFKCCA}, we need to find the IF of $\mb{L}$. All notations and proof are explained in  Appendix. 

It is known that the inverse of an operator may not exits even  exist it may not be continuous operator in general \citep{Fukumizu-SCKCCA}. While we can derive kernel canonical correlation using correlation operator 
$ \vc{V}_{YX}=  \Sigma_{YY}^{- \frac{1}{2}}\Sigma_{YX} \Sigma_{XX}^{- \frac{1}{2}}$, even when $\Sigma_{XX}^{- \frac{1}{2}}$ and  $\Sigma_{YY}^{- \frac{1}{2}}$ are not proper operators, the IF of covariance operator is true only for the finite dimensional RKHSs. For infinite dimensional RKHSs,  we can find IF of $\Sigma_{XX}^{- \frac{1}{2}}$  by  introducing  a regularization term as follows
\begin{multline}
\rm{IF}(\cdot, X^\prime, 
(\Sigma_{XX} + \kappa\vc{I})^{- \frac{1}{2}}) = \frac{1}{2} [(\Sigma_{XX}+\kappa\vc{I})^{- \frac{1}{2}}- (\Sigma_{XX}+\kappa\vc{I})^{- \frac{1}{2}}\tilde{k}_X(\cdot, X^\prime)\otimes  \tilde{k}_X(\cdot, X^\prime)(\Sigma_{XX} + \kappa\vc{I})^{- \frac{1}{2}}], \nonumber
\end{multline}
where $\kappa > 0$ is a regularization coefficient, which gives  empirical estimator.  

Let $(X_i, Y_i)_{i=1}^n$ be a sample from the distribution $F_{XY}$. The EIF of kernel CC and  kernel CV  at $Z^\prime=(X^\prime, Y^\prime)$ for all points $Z_i = (X_i, Y_i)$ are $\rm{EIF} (Z_i, Z^\prime, \rho_j^2) = \widehat{\rm{IF}} (Z^\prime,  \hat{\rho}_j^2),
\rm{EIF} (Z_i, Z^\prime, f_{jX}) = \widehat{\rm{IF}} (\cdot, Z^\prime, f_{jX}), 
 \rm{EIF} (Z_i, Z^\prime, f_{jY}) = \widehat{\rm{EIF}} (\cdot, Z^\prime, \widehat{f}_{jY})$, respectively. 

For the bounded kernels the IFs or EIFs, which are stated in Theorem \ref{TIFKCCA} and after that,  have the three properties: gross error sensitivity, local shift sensitivity and rejection point. But for unbounded kernels, say  a linear, polynomial, the IFs are not bounded. In this consequence, the results of classical kernel CCA using the bounded kernels are  less sensitive than  the results of classical kernel CCA using the unbounded kernels.  In practice, classical kernel CCA affected by the contaminated data even using the  bounded kernels \citep{Ashad-10}.

\subsection{Robust kernel CCA}
\label{sec:RKCCA}
In this Section, we propose  a {\em robust kernel CCA} methods based on robust kernel CO and robust kernel CCO. While many robust linear CCA  methods have  proposed to emphasize  on the linear CCA methods that they fit the bulk of the data well and indicate the points deviating from the original pattern  for further investment \citep{Adrover-15, Ashad-10}, there is no general well-founded robust methods of  kernel CCA. The  classical kernel CCA  considers the same weights for each data point, $\frac{1}{n}$,  to estimate kernel CO and kernel CCO, which is the solution of an empirical risk optimization problem using the quadratic loss function. It is known that the least square loss function is a no robust loss function. Instead of, we can solve  empirical risk optimization problem using the robust least square loss function and  the weights are  determined based on data via KIRWLS. After getting robust kernel CO and kernel CCO, they are used in classical kernel CCA, which we called a {\em robust kernel CCA} method. Figure \ref{RobustKCCA} presents  detailed algorithm  of the proposed methods (except first two steps, all steps are similar as classical kernel CCA). This method  is designed for  contaminated data as well, and  the principles we describe apply also to the  kernel methods, which must deal with the issue of kernel CO and kernel CCO.

\begin{figure}
\hrule
Input: $D=\{(X_1,Y_1), (X_2,Y_2), \ldots, (X_n,Y_n) \}$ in $\mb{R}^{m_1\times m_2}$. 
\begin{enumerate}
\item Calculate the robust cross-covariance operator, $\hat{\Sigma}_{YX}$ using algorithm in Figure \ref{Robust.cross.covariance.M}.
\item Calculate the robust covariance operator $\hat{\Sigma}_{XX}$ and $\hat{\Sigma}_{YY}$ using the same weight of cross-covariance operator (for simplicity).
\item Find $ \mb{B}_{YX} = (\hat{\Sigma}_{YY} + \kappa \vc{I})^{- \frac{1}{2}} \hat{\Sigma}_{YX} (\hat{\Sigma}_{XX} + \kappa\vc{I})^{- \frac{1}{2}}$
\item For $\kappa > 0$, we have $\rho^2_j$  the largest eigenvalue of $\mb{B}_{YX}$ for $j=1, 2,\cdots, n$.
\item  The unit  eigenfunctions of   $\mb{B}_{YX}$  corresponding to the  $j$th eigenvalues  are $\hat{\xi}_{jX}\in \mc{H}_X$ and $\hat{\xi}{j_Y}\in \mc{H}_Y$ 
\item The \it {j}th ($j= 1, 2, \cdots, n$)  kernel canonical variates are given by
\[ \hat{f}_{jX}(X) = \langle \hat{f}_{jX}, \tilde{k}_X(\cdot, X)\rangle \,\rm{and}\,\hat{f}_{jY}(X) = \langle \hat{f}_{jY}, \tilde{k}_Y(\cdot, Y) \rangle\]
 where   $\hat{f}_{jX} =  (\hat{\Sigma}_{XX} + \kappa\vc{I})^{-\frac{1}{2}}\hat{\xi}_{jX}$ and   $f_{jY} =  (\hat{\Sigma}_{YY} + \kappa\vc{I})^{-\frac{1}{2}}\hat{\xi}_{jY}$
\end{enumerate}
Output:  the robust kernel CCA
\hrule
\caption{The algorithm of estimating robust  kernel CCA}
\label{RobustKCCA}
\end{figure}

\section{Experiments}
We generate two types of simulated  data, original data  and those with $5\%$ of  contamination, which are called ideal data (ID) and  contaminated data (CD), respectively.  We conduct experiments on the synthetic data as well as real data sets. The description of $7$ real data sets are in Sections \ref{Sec:Visu} and  \ref{sec:classification}, respectively.  The $5$ synthetic data sets are as follows:

\textbf {Three circles  structural data (TCSD):} 
Data are generated along three circles of different radii with small noise:
\begin{equation}
X_i = r_i \begin{pmatrix} \cos(\vc{Z}_i) \\ \sin(\vc{Z}_i) \end{pmatrix}
+ \epsilon_i, \nonumber
\end{equation}
where $r_i=1$, $0.5$ and $0.25$, for $i=1,\ldots,n_1$, $i=n_1+1,\ldots,n_2$, and $i=n_2+1,\ldots, n3$, respectively, $\vc{Z}_{i}\sim U[-\pi, \pi]$ and $\epsilon_i\sim \mc{N} (0, 0.01\,I_2)$ independently for an ID and  $\vc{Z}_{i}\sim U[-10, 10]$ for the CD. 

\textbf {Sign function structural data (SFSD):}
 1500 data are generated along sine function with small noise:
\begin{equation}
X_i = \begin{pmatrix} \vc{Z}_i \\ 2\sin(2\vc{Z}_i)\\
\vdots \\10\sin(10\vc{Z}_i) \end{pmatrix}
+ \epsilon_i,  \nonumber
\end{equation}
where $\vc{Z}_{i}\sim U[-2\pi,0]$ and $\epsilon_i\sim \mc{N} (0, 0.01\,I_{10})$ independently for the ID and $\epsilon_i\sim \mc{N} (0, 10\,I_{10})$ for the CD.

\textbf {Multivariate Gaussian structural data (MGSD): }
Given  multivariate normal data, $\vc{Z}_i\in\mb{R}^{12} \sim \vc{N}(\vc{0},\Sigma)$ ($i= 1, 2, \ldots, n$) where  $\Sigma$ is the same as in \citep{Ashad-15}.  We   divide $\vc{Z}_i$ into two sets of variables ($\vc{Z}_{i1}$,$\vc{Z}_{i2}$), and use the first six variables of $\vc{Z}_i$ as $X$ and perform $\log$  transformation of the absolute value of the remaining  variables ($\log_e(|\vc{Z}_{i2}|))$) as $Y$. For the CD  $\vc{Z}_i\in\mb{R}^{12} \sim \vc{N}(\vc{1},\Sigma)$ ($i= 1,2,\ldots, n$).

\textbf {Sign and cosine function structural data (SCSD):}
 We use uniform   marginal distribution, and transform the data by two periodic $\sin$ and $\cos$ functions to make two sets $X$ and $Y$, respectively, with additive Gaussian noise:
$Z_i\sim U[-\pi,\pi], \,\eta_i\sim N(0,10^{-2}),~\,i=1,2,\ldots, n,
 X_{ij}=\sin(j*Z_i)+\eta_i,\,  Y_{ij} = \cos(j*Z_i)+\eta_i, j=1,2,\ldots, 100.$
For the CD $\eta_i\sim N(1,10^{-2})$.

\textbf {SNP and fMRI structural data (SMSD):}
Two data sets of SNP data X with $1000$ SNPs and fMRI data Y with 1000
voxels were simulated. To correlate the SNPs with the voxels, a  latent
model is used   as in \cite{Parkhomenko-09}). For contamination,  we consider the signal level, $0.5$  and  noise level, $1$  to $10$ and $20$, respectively.

In our experiments, first we compare classical and robust  kernel covariance operators. After that the robust kernel CCA  is compared with the classical kernel CCA, hsicCCA and ktaCCA. In all experiments, for the Gaussian kernel we use  median of the  pairwise distance as a  bandwidth  and  for the Laplacain kernel the bandwidth is equal to 1. The regularized parameters of classical  kernel CCA and robust kernel CCA is  $\kappa= 10^{-5}$. In robust methods, we consider  Hubuer's loss function with the constant, $c$, equals to the median.

\subsection{Kernel covariance operator and robust kernel operator covariance}
We evaluate the performance of kernel CO  and robust kernel CO in two different settings. First, we  check the accuracy of both operators by considering the  kernel  CO  with large data (say a population kernel CO). The measure of the kernel CO and   robust  kernel CO  estimators are  defined as
\begin{multline}
\eta_{kco}=\frac{1}{n^2}\sum_{i=1}^n\sum_{j=1}^nk(X_i,X_j)^2 - 2\frac{1}{Nn}\sum_{i=1}^n\sum_{J=1}^Nk(X_i,X_J)^2+ \frac{1}{N^2}\sum_{I=1}^N\sum_{J=1}^Nk(X_I,X_J)^2,\,\nonumber \rm{and}\end{multline}
\begin{multline}
\eta_{rkco}=\sum_{i=1}^n\sum_{j=1}^n w_iw_jk(X_i,X_j)^2- 2\frac{1}{N}\sum_{i=1}^n w_i\sum_{J=1}^Nk(X_i,X_J)^2 + \frac{1}{N^2}\sum_{I=1}^N\sum_{J=1}^Nk(X_I,X_J)^2, \nonumber 
\end{multline}
 respectively.

\begin{figure}[t] 
\begin{center}
\includegraphics[width=\columnwidth, height=12cm]{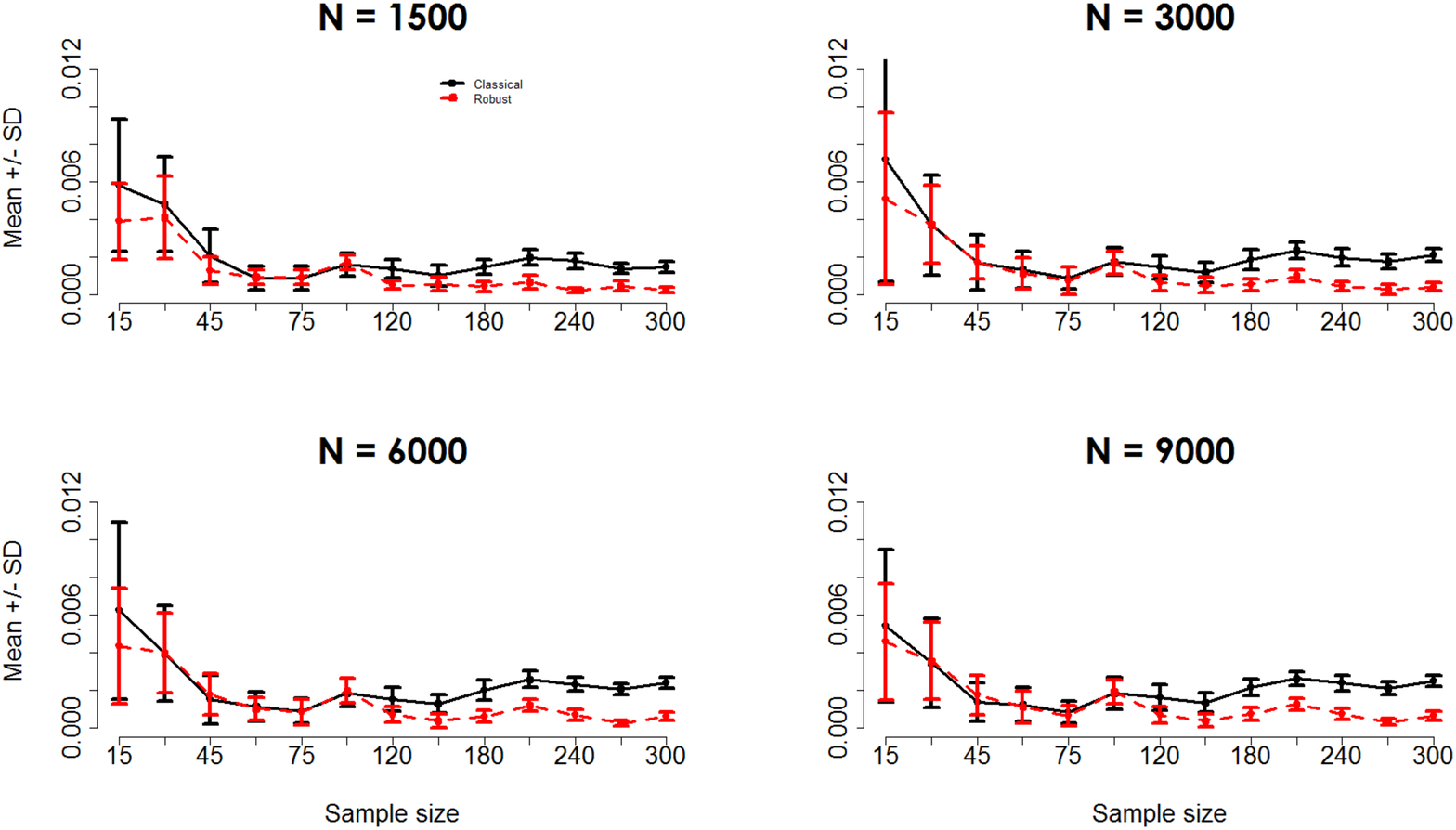}
\caption{Accuracy measure of kernel covariance operator (Classical, $\eta_{kco}$) and robust kernel covariance operator (Robust, $\eta_{rkco}$).}
\label{Fcons}
\end{center}
\end{figure}

In theory, the above two equations become to zero for large population size, $N$, with the sample size, $n\to N$. To do this, we consider the synthetic data, TCSD with  $N\in\{1500, 3000, 6000, 9000\}$ and   $n\in\{15, 30, 45, 60, 90, 120, 150, 180, 210, 240, 270, 300\}$ ($n = n_1 + n_2 + n_3$). For each  $n$ with $5\%$ CD, we  consider  $100$ samples and the results (mean with standard error) are plotted in Figure \ref{Fcons}. Figures show that the both estimators give similar performance in small sample size but for large sample sizes  the   robust estimator, robust kernel CO shows much better results than kernel CO estimate  at all  population sizes.  

Second, we compare kernel CO and robust  kernel CO  estimators using 5 kernels: linear (Poly-1), polynomial with degree $2$ (Poly-2) and polynomial with degree $3$ (Poly-3), Gaussian and Laplacian on two synthetic data sets: TCSD and SFSD. To measure the performance, we use 4 matrix norms:  maximum  of absolute column sum (O), Frobenius norm (F), maximum modulus of all the elements (M) and spectral (S) \citep{Sequeira-11}.   We calculate the ratio  between ID  and CD for the kernel CO  and robust kernel CO. For both estimators,  we consider the following measure, 
\[\eta_{co} = | 1- \frac{\| \hat{C_{XX}}^{ID}\|}{\|\hat{C_{XX}}^{CD} \|} |.\]
 We repeat the experiment  for 100 samples with sample size, $n=1500$. The results (mean $\pm$ standard deviation) of $\eta_{co}$ for kernel CO (Classical) and robust kernel CO (Robust)   are tabulated in  Table \ref{NRCRcov}.  From this table, it is clear that  the  robust estimator  performs better than the classical estimator in all cases. Moreover, both estimators  using   Gaussian and Lapalasian kernels are less sensitive than all   polynomial kernels.   
\begin{table}[t]
\begin{center}
\caption {Mean  and  standard deviation of the  measure, $\eta_{cov}$, of  kernel covariance operator (Classical) and  robust kernel covariance operator (Robust).}
\label{tbl:norm}
 \begin{tabular}{llcccc} \hline
&Data&\multicolumn{2}{c}{TCSD}&\multicolumn{2}{c}{SFSD}\\ \cline{2-6}
Measure&Kerenl&Classical&Robust &Classical&Robust    \\ \hline
&Poly-1&$0.9947\pm 0.0007$&$0.9546\pm 0.0026$&$0.5692\pm 0.1426$&$0.4175\pm 0.1482$\tabularnewline
&Poly-2&$1.0000\pm 0.0000$&$0.9909\pm 0.0013$&$0.9652\pm  0.0494$&$  0.6703\pm  0.172$\tabularnewline
$\| \hat{C}_{XX} \|_O$&Poly-3&$1.0000\pm 0.0000$&$0.9997 \pm 0.0001$&$0.9971\pm  0.0095$&$  0.8754 \pm 0.1104$\tabularnewline
&Gaussian&$0.0844\pm 0.0054$&$0.0756\pm 0.0051$&$0.1167\pm  0.0459$&$  0.0964  0.0424$\tabularnewline
&Laplacian&$0.1133\pm 0.0054$&$0.0980\pm 0.0128$&$0.1332 \pm 0.0830 $&$ 0.1420 \pm 0.0745$\tabularnewline\hline
&Poly-1&$0.9874\pm 0.0017$&$0.8963\pm 0.0069$&$0.3067\pm  0.1026$&$  0.1669 \pm 0.0626$\tabularnewline
&Poly-2&$1.0000\pm 0.0000$&$0.9863\pm 0.0020$&$0.9559\pm 0.0622$&$  0.5917  \pm 0.1598$\tabularnewline
$\| \hat{C}_{XX} \|_F$&Poly-3&$1.0000 \pm 0.0000$&$0.9996\pm 0.0001$&$0.9973\pm 0.0094 $&$ 0.8793  \pm 0.1067$\tabularnewline
&Gaussian&$0.1153\pm 0.0034$&$0.1181\pm 0.0039$&$0.1174\pm 0.0266 $&$ 0.1059  \pm 0.0258$\tabularnewline
&Laplacian&$0.1420\pm 0.0032$&$0.1392\pm 0.0035$&$0.1351\pm 0.0459 $&$ 0.1580  \pm 0.0366$\tabularnewline \hline
&Poly-1&$0.9993\pm 0.0001$&$0.9940\pm 0.0005$& $0.8074 \pm 0.0838$&$  0.6944\pm  0.1118$\tabularnewline
&Poly-2&$1.0000\pm 0.0000$&$0.9996\pm 0.0001$&$0.9921 \pm  0.0122 $&$ 0.9070 \pm 0.0703$\tabularnewline
$\| \hat{C}_{XX} \|_M$&Poly-3&$1.0000\pm 0.0000$&$1.0000\pm 0.0000$&$0.9994 \pm 0.0020 $&$ 0.9709 \pm 0.0344$\tabularnewline
&Gaussian&$0.1300\pm 0.0133$&$0.1028\pm 0.0038$&$0.1065 \pm 0.0583 $&$  0.0735 \pm 0.0370$\tabularnewline
&Laplacian&$0.1877\pm 0.0053$&$0.1474\pm 0.0042$&$0.1065 \pm 0.0583 $&$  0.0735 \pm 0.0370$\tabularnewline
\hline
&Poly-1&$0.9886\pm 0.0019$&$0.9007\pm 0.0091$&$0.2897\pm  0.1412 $&$ 0.1538\pm 0.0877$\tabularnewline
&Poly-2&$1.0000\pm 0.0000$&$0.9887\pm 0.0017$&$0.9591 \pm 0.0660 $&$ 0.5927 \pm 0.2002$\tabularnewline
$\| \hat{C}_{XX} \|_S$&Poly-3&$1.0000\pm 0.0000$&$0.9997\pm 0.0001$&$0.9975 \pm 0.0090 $&$ 0.8846 \pm 0.1225$\tabularnewline
&Gaussian&$0.1281\pm 0.0051$&$0.1227\pm 0.0048$&$0.1333 \pm 0.0475 $&$ 0.1091 \pm 0.0459$\tabularnewline
&Laplacian&$0.1716\pm 0.0050$&$0.1499\pm 0.0045$&$0.1614 \pm 0.0759 $&$ 0.1696 \pm 0.0613$\tabularnewline
\hline
\end{tabular}
\label{NRCRcov}
\end{center}
\end{table}

\subsection{Visualizing influential subject using classical kernel CCA and robust kernel CCA}
\label{Sec:Visu}
We evaluate the performance of  the propose methods, {\em robust kernel CCA}, in three different settings. First, we compare  robust kernel CCA with the classical kernel CCA using Gaussian kernel (same bandwidth and regularization). To measure the influence, we calculate the  ratio between  ID and CD of IF of kernel CC and kernel CV. Based on this ratio, we define two  measure  on kernel  CC and kernel CV  
\begin{eqnarray}
\eta_{\rho} &=& | 1- \frac{\|EIF(\cdot, \rho^2) ^{ID}\|_F}{\|EIF(\cdot, \rho^2)^{CD}\|_F}| \qquad \rm{and} \\ 
\eta_{f} &=& | 1- \frac{\|EIF(\cdot, f_X)^{ID}- EIF(\cdot,f_Y)^{ID}\|_F}{\|EIF(\cdot, f_X) ^{CD}-EIF(\cdot, f_Y)^{CD}\|_F}|,\nonumber
\end{eqnarray}
respectively. The method, which does not depend on the contaminated data, the above measures, $\eta_{\rho}$  and $\eta_{f}$, should  be approximately zero. In other words, the best methods should give small values. To compare, we consider 3 simulated data sets:  MGSD, SCSD, SMSD with 3 sample sizes, $n\in \{ 100, 500, 1000\}$. For each sample size, we repeat the experiment for $100$ samples.  Table \ref{tbl:ifnorm} presents the results (mean $\pm$ standard deviation) of classical kernel CCA and robust kernel CCA. From this table, we observe that robust kernel CCA outperforms than the classical kernel CCA in all cases.  

\begin{table}[t]
 \begin{center}
\caption {Mean and standard deviation of the measures, $\eta_{\rho}$ and $\eta_{f}$ of classical kernel CCA (Classical) and  robust kernel CCA (Robust).}
\label{tbl:ifnorm}
 \begin{tabular}{llcccccccc} \hline
&Measure &\multicolumn{2}{c}{$\eta_{\rho}$}&\multicolumn{2}{c}{$\eta_{f}$}\\ \cline{3-6}
Data&n&$Classical$&$Robust$&$Classical$&$Robust$  \\ \hline
&$100$&$1.9114\pm 3.5945$&$ 1.2445\pm 3.1262$&$ 1.3379\pm 3.5092$&$ 1.3043\pm 2.1842$\tabularnewline
MGSD&$500$&$ 1.1365\pm 1.9545$&$ 1.0864\pm 1.5963$&$ 0.8631 \pm 1.3324 $&$0.7096\pm 0.7463$\tabularnewline
&$1000$&$ 1.1695\pm 1.6264$&$ 1.0831\pm 1.8842$&$ 0.6193 \pm 0.7838$&$ 0.5886\pm 0.6212$\\ \hline
&$100$&$0.4945\pm 0.5750$& $0.3963\pm 0.4642$& $1.6855\pm 2.1862$& $0.9953\pm 1.3497$\tabularnewline
SCSD &$500$&$0.2581\pm 0.2101$& $0.2786\pm 0.4315$& $1.3933\pm 1.9546$&$ 1.1606\pm 1.3400$\tabularnewline
&$1000$&$0.1537\pm 0.1272$&$ 0.1501\pm 0.1252$&$ 1.6822\pm 2.2284$&$1.2715\pm 1.7100$\\ \hline
&$100$&$ 0.6455 \pm 0.0532$& $ 0.1485\pm 0.1020$& $ 0.6507 \pm 0.2589 $& $2.6174\pm 3.3295$ \tabularnewline
SMSD&$500$&$0.6449\pm 0.0223$& $0.0551\pm 0.0463$&$ 3.7345 \pm 2.2394$& $ 1.3733 \pm 1.3765$\tabularnewline
&$1000$&$ 0.6425 \pm 0.0134$& $ 0.0350\pm 0.0312$& $ 7.7497\pm 1.2857$& $ 0.3811 \pm 0.3846$\\ \hline
\end{tabular}
\end{center}
\end{table}

Second, we propose a simple graphical display based on EIF of kernel CCA, the index plots (the subject on $x$-axis and the influence, $\eta_{\rho}$, on $y$ axis), to assess the related influence data points in data fusion with respect to EIF based on kernel CCA, $\eta_{\rho}$. To do this, we consider simulated  SNP and fMRI data (SMSD) and real SNP and fMRI, Mind Clinical Imaging Consortium (MCIC) Data. 
\begin{figure}[t] 
\begin{center}
\includegraphics[width=\columnwidth, height=12cm]{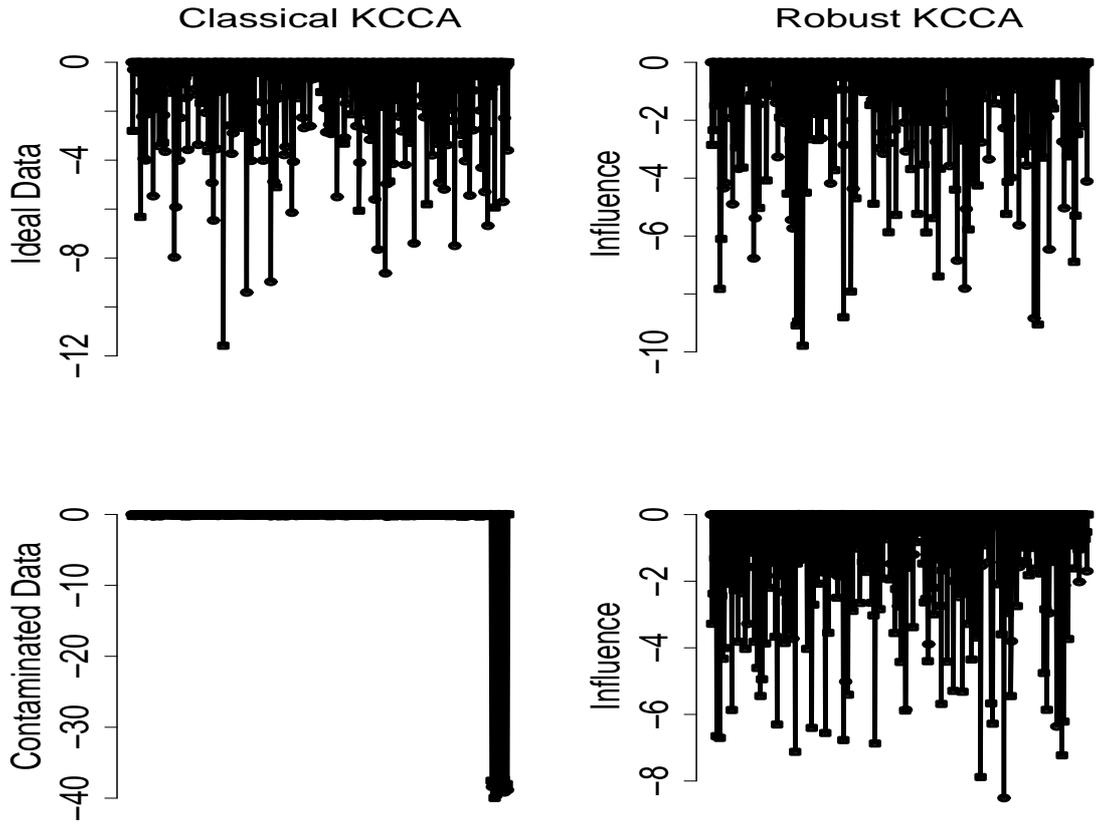}
\caption{Influence  points of SMDS   for ideal data and  contaminated data using classical and robust kernel CCA.}
\label{SMDSIFOB}
\end{center}
\end{figure}

{\bf Mind Clinical Imaging Consortium (MCIC) Data:} 
The Mind Clinical Imaging Consortium (MCIC) has collected two types of data (SNPs and fMRI) from 208 subjects including $92$ schizophrenia patients and $116$ healthy controls. Without missing data the number of subjects is $184$ ($81$ schizophrenia patients and $103$ healthy controls) \citep{Dongdong-14}. After prepossessing  we select $19872$ voxels and  $39770$ loci for fMRI data and SNP data, respectively.

The index plots of  classical kernel CCA and robust kernel CCA using the SMSD  are presented in Figure \ref{SMDSIFOB}. The $1$st and $2$nd rows, and columns of this figure are  for ID and CD, and classical kernel CCA (Classical KCCA) and robust kernel CCA (Robust KCCA), respectively.  These plots show that both methods have almost similar results of  the ID. But for CD, it is clear that the classical kernel CCA is affected by the CD in significantly. We can easily identify influence of observation  using this visualization. On the other hand, the robust kernel CCA has almost similar results of both data sets, ID and CD. 

\begin{figure}[t] 
\begin{center}
\includegraphics[width=\columnwidth, height=12cm]{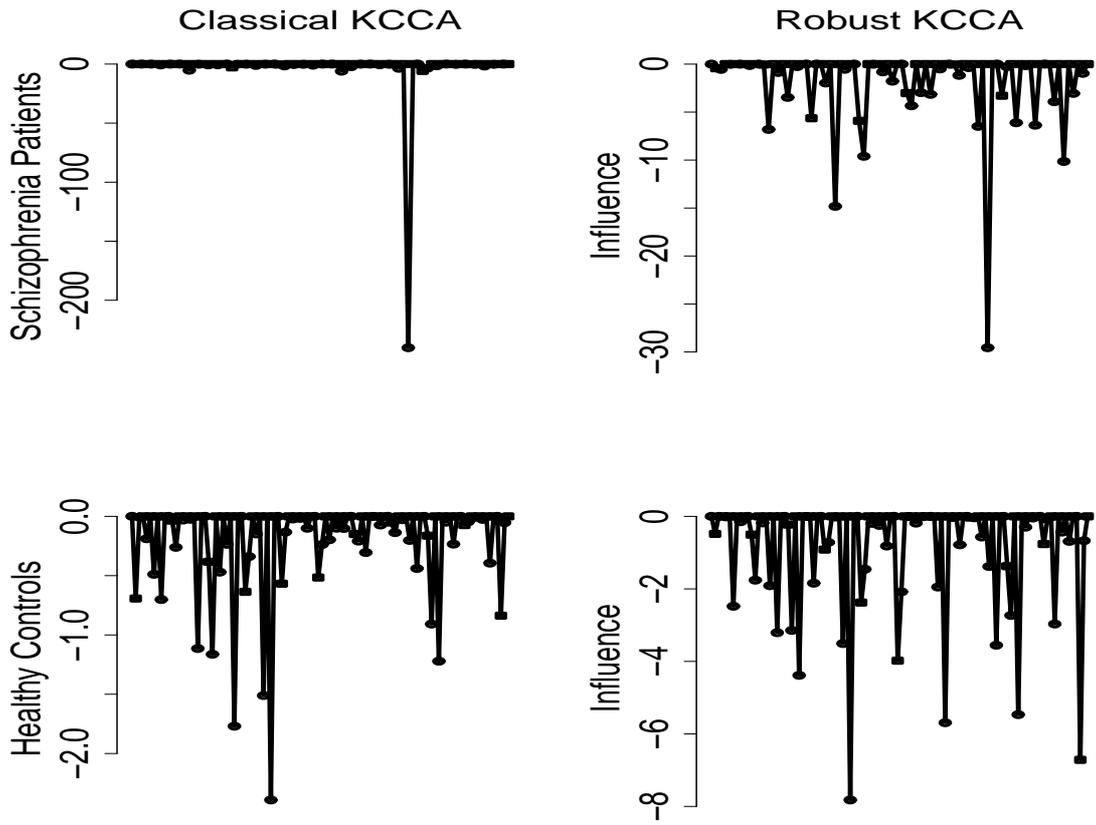}
\caption{Influence  subject of MCIC data set of classical and robust KCCA of the  real data.}
\label{MCICFOB}
\end{center}
\end{figure}

To detect influential subjects (in schizophrenia patients and healthy controls), we use the EIF of kernel CC of  classical and robust kernel CCA methods. For robust  kernel CCA, we use robust kernel CC and kernel CVs in Theorem \ref{TIFKCCA}.   The values of  $\eta_{\rho}$  are plotted  separately in  Figure \ref{MCICFOB}. The   schizophrenia patients and  healthy controls  are in $1$st and $2$nd rows, respectively. These plots show that  healthy controls  have less influence than the   schizophrenia patients group.  The subject $59$ in  Schizophrenia patients has the largest influence over all data and the subject $119$  has the largest influence over healthy controls only. However, both  classical and robust kernel CCA have identified similar subject but robust kernel CCA  is less sensitive than classical kernel CCA.

\subsection{Extraction of low-dimensional space for classification}
\label{sec:classification}
Finally, we use  $6$  well-known real  datasets for classification from the UCI repository \citep{UCI}: {\it Wine}, {\it BUPA liver disorders}, {\it Breast tissue},  {\it Diabetes}, {\it Sona}, and  {\it Lymphoma} to test the significance of  low dimensional canonical features of the input space.  We use the features for the classification task. To specify the classes, for an $\ell$-class classification problem, the $\ell$ dimensional binary vectors $(1, 0,\ldots 0), (0, 1,\ldots 0), \ldots (0, 0,\ldots 1)$ are used for $Y$. 

Using the low-dimensional canonical features (only $2$, $5$ and $10$) obtained by CCA, we evaluate the classification errors by the kNN classifier ($k = 5$) with $10$-fold cross-validation. In comparison, we use the canonical features given by the classical kernel CCA, robust kernel CCA, hsicCCA and ktaCCA methods. For  hsicCCA and ktaCCA methods we use "hsicCCA" R-package.  The Table \ref{tb3:clasification} presents the results with the number of data and dimensions.  By this table, we see that the  hsicCCA and ktaCCA methods are not able to extract  all of the canonical variates. On top of that, these methods do not work  for the high dimensional dataset. In  Table \ref{tb3:clasification},   these situation are noted by $*$ and $**$ respectively. On the other hand, classical kernel CCA and robust kernel CCA can extract  all canonical variates as well as for  all data sets. In fact, the kernel CCA methods are more faster than the  hsicCCA and ktaCCA methods.  
\begin{table}[t]
 \begin{center}
\begin{threeparttable}
\caption {Classification errors by the kNN classifier ($k = 5$) with $10$-fold cross-validation}
\label{tb3:clasification}
 \begin{tabular}{lllcccccccc} \hline
&&\multicolumn{2}{c}{\rm{Dim}}&&\multicolumn{3}{c}{\rm{ \# of canonical variates}}\\ \cline{3-4}\cline{6-8}
\rm{Dataset}&\# \rm{of Data}&X&Y&\rm{Methods}&$2$&$5$&$10$\tabularnewline  \hline
&$$&$$&$$&Classical KCCA &$ 2.81$&$0.56$&$0.56$\tabularnewline
Wine&$178$&$13$&$3$&Robust KCCA&$2.81$&$0.56$&$0.56$\tabularnewline
&$$&$$&$$&hsicCCA&$2.47$&$*$&$*$\tabularnewline
&$$&$$&$$&ktaCCA&$2.47$&$*$&$*$\tabularnewline\hline
&$$&$$&$$&Classical kernel CCA &$ 36.79$&$20.76$&$17.92$\tabularnewline
Breast Tisu&$106$&$10$&$6$&Robust kernel CCA&$38.68$&$19.81$&$18.81$\tabularnewline
&$$&$$&$$&hsicCCA&$ 24.53$&$20.75$&$*$\tabularnewline
&$$&$$&$$&ktaCCA&$22.64$&$20.75$&$*$\tabularnewline \hline
&$$&$$&$$&Classical kernel CCA &$12.41$&$5.52$&$4.14$\tabularnewline
Diabetes&$145$&$5$&$3$&Robust kernel CCA&$12.41$&$5.52$&$4.14$\tabularnewline
&$$&$$&$$&hsicCCA&$ 11.03$&$*\tnote{}$&$*$\tabularnewline
&$$&$$&$$&ktaCCA&$11.03$&$*$&$*$\tabularnewline \hline
&$$&$$&$$&Classical kernel CCA &$14.90$&$14.90$&$13.94$\tabularnewline
Sona&$208$&$60$&$2$&Robust kernel CCA&$13.94$&$13.94$&$13.94$\tabularnewline
&$$&$$&$$&hsicCCA&$**\tnote{}$&$**$&$**$\tabularnewline
&$$&$$&$$&ktaCCA&$**$&$** $&$**$\tabularnewline \hline
&$$&$$&$$&Classical kernel CCA &$1.61$&$0.00$&$0.00$\tabularnewline
Lymphoma&$64$&$4026$&$3$&Robust kernel CCA&$1.61$&$0.00$&$0.00$\tabularnewline
&$$&$$&$$&hsicCCA&$**$&$**$&$**$\tabularnewline
&$$&$$&$$&ktaCCA&$**$&$**$&$**$\tabularnewline \hline
\end{tabular}
\end{threeparttable}
\end{center}
\begin{tablenotes}
            \item[a]*Functions cannot be evaluated at initial parameters
            \item[b]**Curse  of dimensionality
        \end{tablenotes}
\end{table}

\section{Concluding remark and future research}
The robust estimator employs robust loss function instead of quadratic loss function  to achieve robustness for the contamination of the training sample. The robust estimators are  weighted estimators, where smaller weights are given more outlying data points. The weights can be estimated efficiently  using  a KIRLS approach. In terms of  the accuracy and sensitivity, it is clear that the robust  estimators (robust kernel CO and robust kernel CCO) perform better than classical estimators (kernel CO and kernel CCO). We propose the IF of  kernel CCA: kernel CC and kernel CVs and  {\em robust kernel CCA} based on robust kernel CO and robust kernel CCO.  The proposed IF measures  the sensitivity of kernel CCA, which shows that classical kernel CCA is sensitive to contamination. But the proposed, {\em robust kernel CCA} is less sensitive to contamination. The visualization  methods can identify influential (outlier) data in both synthesized and  real imaging genetics analysis data. We also  obtain low dimensional subspace for classification  by CCA.  we evaluate the classification errors by the kNN classifier ($k = 5$) with $10$-fold cross-validation.  The proposed {\em robust  CCA}  shows the best performance over   hsicCCA and ktaCCA methods.

For the EIF of  robust  kernel CCA, we use robust kernel CC and kernel CVs in Theorem \ref{TIFKCCA}. The theoretical IF of robust kernel CCA is an expected future direction of research. Although our focus was on kernel CCA but we can robustify other kernel methods, which must deal with the issue of  kernel CO and  kernel CCO. In future work, it would be also interesting to develop robust multiple kernel PCA and robust multiple weighted kernel CCA to apply in genomic analysis. 
\begingroup
\bibliographystyle{plainnat}
\bibliography{Ref-UKIF}
\endgroup
\section{Appendix}

We present, derivation of   robust centering Gram matrix, robust kernel cross-covariance operator,  influence function (IF) of kernel mean elements and kernel cross-covariance operator and  proofs which were omitted from the paper.

\subsection{Derivation of centering  Gram matrix using robust kernel mean element}
Given weight of robust kernel mean element $\vc{w}=[ w_1, w_2, \cdots, w_n]^T$ of a set of observations $X_1, \cdots, X_n$, the points 
\begin{eqnarray}
\vc{\Phi}_c(X_i):= \vc{\Phi}(X_i)- \sum_{b=1}^nw_b \vc{\Phi}(X_b) \nonumber
\end{eqnarray} 
 are centered. Thus
\begin{eqnarray}
\tilde{K}_{ij}&=&\langle \vc{\Phi}_c(X_i), \vc{\Phi}_c(X_j)\rangle   \nonumber\\
&=& \left\langle \vc{\Phi}(X_i) - \sum_{b=1}^nw_b  \vc{\Phi}(X_b),  \vc{\Phi}(X_j)- \sum_{d=1}^nw_d  \vc{\Phi}(X_d) \right\rangle  \nonumber \\
&=&\langle \vc{\Phi}(X_i),  \vc{\Phi}(X_j)\rangle - \sum_{b=1}^nw_b  \langle \vc{\Phi}(X_b) , \vc{\Phi}(X_j)\rangle - \sum_{d=1}^nw_d  \langle \vc{\Phi}(X_i) , \vc{\Phi}(X_d)\rangle + \sum_{b=1}^n \sum_{d=1}^nw_b w_d  \langle \vc{\Phi}(X_b), \vc{\Phi}(X_d)\rangle \nonumber\\
&=&\vc{K}_{ij} - \sum_{b=1}^nw_bK_{bj} - \sum_{d=1} K_{id}w_d+\sum_{b=1}^n\sum_{d=1}^n w_bK_{bd}w_d  \nonumber\\
&=&(\vc{K}- \vc{1}_n \vc{w}^T \vc{K}- \vc{K} \vc{w}\vc{1}_n^T+ \vc{1}_n \vc{w}^T \vc{K} \vc{w}\vc{1}^T_n)_{ij} \nonumber \\
&=& ( (\vc{I}- \vc{1}_n\vc{w}^T) \vc{K} (\vc{I}- \vc{1}_n \vc{w}^T)^T)_{ij} \nonumber\\
&=&(\vc{H}\vc{K}\vc{H}^T)_{ij},
\label{CM1}
\end{eqnarray}
where $\vc{1}_n=[1, 1, \cdots, 1]^T$ and $\vc{H}=\vc{I}- \vc{1}_n\vc{w}^T$. For a set of test points $ X^t_1, X_2^t, \cdots, X_T^t$,  we define two matrices of order $T\times n$ as  $K_{ij}^{test}= \langle \Phi(X^t_i), \Phi(X_j) \rangle$ and  $\tilde{K}_{ij}^{test}= \langle \Phi(X^t_i)- \sum_{b=1}^nw_b \Phi(X_b),  \Phi(X_j)- \sum_{d=1}^nw_d  \Phi(\vc{X}_b)\rangle$
As in Eq. (\ref{CM1}) the robust centered Gram matrix of test points,  $K_{ij}^{test}$,  in terms of  robust Gram matrix is defined as,
\begin{eqnarray}
\label{CM2}
\tilde{K}_{ij}^{test}=  K_{ij}^{test}- \vc{1}_T \vc{w}^T \vc{K}- \vc{K}^{test} \vc{w}\vc{1}_n^T+ \vc{1}_T \vc{w}^T \vc{K} \vc{w}\vc{1}^T_n \nonumber
\end{eqnarray}

\subsection{Derivation of centering  Gram matrix using robust kernel mean element}

Similarly, we can define higher-order moment elements  of the feature vector as  an empirical risk optimization problem.
\begin{defn} [Kernel  \textit{r}th raw moment element]
A mapping $\mc{M}^{(r)}:= \mb{E}_X[\Phi(X)\otimes \Phi(X)\otimes \cdots \otimes \Phi(X)] = \otimes^r\Phi(X)=\mb{E}_X[k(X, \cdot)k(X, \cdot)\cdots k(X, \cdot)]= \mb{E}_X[k^{(r)}(\cdot, X)]$ with $\mb{E}_X[k^{(r)}(X, X)] < \infty$ is an element of the RKHS,  $\otimes^r\mc{H}_X$.  By the reproducing property  with $X\in \mc{X}$
\begin{eqnarray}
\langle \mc{M}^{(r)}, \otimes^r f \rangle_{\otimes^r\mc{H}_X} =  \mb{E}_X[\langle \mc{M}, f \rangle_{\mc{H}_X} \langle \mc{M}, f \rangle_{\mc{H}_X} \cdots \langle \mc{M}, f \rangle_{\mc{H}_X}]
=\mb{E}_X[f(X) f(X)\cdots f(X)] 
=  \mb{E}_X[f^{(r)}(X)],  
\end{eqnarray}
 for all  $f\in \mc{H}_X$. 
 The  mapping   $m^{(r)}= \frac{1}{n}\sum_{i=1}^n \otimes^r \Phi(X_i)$ is an   empirical \it{r}th row moment element of the RKHS,  $\otimes^r\mc{H}_X$, 
\[\langle  m^{(r)}, \otimes^r f\rangle_{\otimes^r\mc{H}_X}= \frac{1}{n}\sum_{i=1}^nf^{(r)}(X_i),\]
 where $\otimes^r f= f\otimes f \otimes \cdots \otimes f$ is the tensor product of $r$ functions, $f \in \mc{H}_X$.
The sample \it{r}th  row moment  element of the $ \vc{\Phi}(X_i) $ is a solution  of  an empirical risk optimization problem
\begin{eqnarray}
\label{rmop} 
\argmin_{g\in \otimes^r\mc{H}_X} \frac{1}{n}\sum_{i=1}^n\| \otimes_{j=1}^r\vc{\Phi}^{(r)}(X_i) - g\|^2_{\otimes^r\mc{H}_X}, 
\end{eqnarray}
 at the point $X$,  $g(X, X,\cdots X) \in \otimes^r\mc{H}_X$. 
\end{defn}

\begin{defn}[Kernel  \textit{r}th central moment element]
\label{KCME}
A mapping $\mc{M}_c^{(r)}:= \mb{E}_X[\tilde{k}(\cdot,X)\tilde{k}(\cdot, X)\cdots \tilde{k}(\cdot, X)]$ with $\mb{E}_X[\tilde{k}^{(r)}(X, X)] < \infty$ is an element of the RKHS,  $\otimes\mc{H}_X$. By the reproducing property   $\forall k(\cdot, X),\,  f\in \mc{H}_X\, \rm{and}\,, X\in \mc{X}\, \forall f\in \mc{H}_X$,
\[
\langle  \mc{M}_c^{(r)} , \otimes^r f_c \rangle_{\otimes\mc{H}_X}= \langle \mc{M}_c ,f_c \rangle_{\mc{H}_X} \langle \mc{M}, f_c \rangle_{\mc{H}_X} \cdots \langle \mc{M}_c[F_X] ,f_c \rangle_{\mc{H}_X}=  \mb{E}_X[k_c^{(r)}(\cdot, X)] \]
and the empirical \textit{r}th central moment element at every point $X_i$ is defined by
\[
\langle \mc{M}_c^{(r)}, \otimes^r f_c\rangle_{\otimes\mc{H}_X}= \frac{1}{n}\sum_{b=1}^nf_c^{(r)}(X_b)=  \frac{1}{n}\sum_{b=1}^nk_c^{(r)}(X_i, X_b).\]

The sample \it{r}th kernel moment  element of the $ \vc{\Phi}(X_i) $ is a solution  of 
\begin{eqnarray}
\label{cmop}
\argmin_{\otimes f_c\in \otimes\mc{H}_X} \frac{1}{n}\sum_{i=1}^n\| \otimes_{j=1}^r\vc{\Phi}_c^{(r)}(X_i)- \otimes^r f_c\|^2_{\otimes\mc{H}_X}\nonumber\\
 =\argmin_{\otimes g_c\in \otimes\mc{H}_X} \frac{1}{n}\sum_{i=1}^n\| \otimes_{j=1}^r\vc{\Phi}_c^{(r)}(X_i)- g\|^2_{\otimes\mc{H}_X}
\end{eqnarray}
where $\vc{\Phi}_c^{(r)}(X_i)= \vc{\Phi}^{(r)}(X_i)-\frac{1}{n}\sum_{i=1}^n \vc{\Phi}^{(r)}(X_i)$, at point  $X$ , $g_c(X, X,\cdots X)= f_c(X) f_c(X) \cdots f_c(X)=f_c^{(r)}(X)$  and   $\otimes f_c = f_c\otimes f_c \otimes \cdots \otimes f_c$ is the tensor product of $r$ functions $f_c \in \mc{H}_X$.  
\end{defn}

\subsection{Influence function of  mean element and cross-raw moment}

\begin{defn} ({\bf Influence function}). Let ($\Omega$, $\mc{A}$, $\mb{P}$) be a probability space and $(\mc{X}, \mc{B})$  a measure space. We want to estimate the parameter $\theta \in \Theta$  of a distribution  $F$ in $\mc{A}$.  We assume that exists a functional $R: \mc{D}(R) \to \mb{R}$, where $ \mc{D}(R)$  is the set of all  probability distribution in $\mc{A}$.
Let $G$ be some distribution in $\mc{A}$. If data do not fallow the model $F$ exactly but  slightly  going toward $G$,  the  G\^{a}teaux Derivative at $F$ is given by 
\begin{eqnarray}
\lim_{\epsilon\to 0 }\frac{R[(1-\epsilon)F+\epsilon G] - R(F)}{\epsilon}
\end{eqnarray}
Suppose $x\in \mc{X}$ and $G=\Delta_x$ is the probability measure which gives mass $1$  to $\{x\}$. The influence function (special case of G\^{a}teaux Derivative) of $R$ at $F$ is defined by

\begin{eqnarray}
IF(x, R, F)=\lim_{\epsilon\to 0 }\frac{R[(1-\epsilon)F+\epsilon\Delta_x] - R(F)}{\epsilon}
\end{eqnarray}
provided that the limit exists. It can be intuitively interpreted as a suitably normalized asymptotic influence of outliers on the value of an estimate or test statistic.
\end{defn}
The equivalent definition can also be defined using the perturbation theory. Consider the case where $R(\epsilon) = R[(1-\epsilon)F+\epsilon G] - R(F)$ is expanded as a convergent power series of $\epsilon$ as
\begin{eqnarray}
R(\epsilon)= R + \epsilon R^{(1)}+ \epsilon^2 R^{(2)}+ O(\epsilon^3) \nonumber 
\end{eqnarray} 
 Due to  the properties of convergent power series $R(\epsilon)$  is differentiable  in a neighborhood of $\epsilon=0$. The IF,  $IF(x, R)$  equals to $ R^{(1)}$, the first order term of $\epsilon$. There are three properties of IF: gross error sensitivity, local shift sensitivity and  rejection point.  They  measured the worst effect of gross error, the worst effect of rounding error and  rejection point.

For a scalar, we just define influence function (IF) at a fixed point. But if the estimate is a function, we are able to express the change of the function value at every points \citep{Kim-12}.
\begin{Exm} [Kernel  mean element] Let 
$R(F_X)=\langle \mc{M}, f \rangle_{\mc{H}_X}= \mb{E}_X[f(X)]= \int f(X)dF_X=  \int k(\cdot, X)dF_X.$  The value of parameter at the  contamination model, $W^\epsilon= (1-\epsilon)F_X+\epsilon\Delta_{X^\prime}$ is given by
\begin{eqnarray}
R\left[W^\epsilon\right]=R\left[(1-\epsilon)F_X+\epsilon\Delta_{X^\prime}\right]&=& \int f(\tilde{X})d[(1-\epsilon)F_X+\epsilon \Delta_{X^\prime}] \nonumber \\
&=&(1-\epsilon)\int f(\tilde{X})dF_X+\epsilon\int f(\tilde{X})d_{\Delta_{X^\prime}}) \nonumber \\
&=&(1-\epsilon)\int k(\tilde{X},X)dF_X+\epsilon\int  k(\tilde{X}, X)d_{\Delta_{X^\prime}} \nonumber \\
&=& (1-\epsilon)\int  k(X, \tilde{X})dF_X +\epsilon  k(\tilde{X},X^\prime)\nonumber\\
&=& (1-\epsilon) R(F_X) + \epsilon  k(\tilde{X},X^\prime) \nonumber
\end{eqnarray}
Thus the IF of $R(F_X)$ at point $X^\prime$ for every point $\tilde{X}$ is given by
\begin{eqnarray}
IF(X^\prime, \tilde{X}, R, F_X) &=&\lim_{\epsilon\to 0 }\frac{R[W_X^\epsilon] - R(F_X)}{\epsilon} \nonumber\\
&=&\lim_{\epsilon\to 0 }\frac{(1-\epsilon)RF_X+\epsilon  k( \tilde{X},X^\prime)-R(F_X)]}{\epsilon} \nonumber\\
&=&\lim_{\epsilon\to 0 }\biggl[k(\tilde{X},X^\prime) - R(F_X) \biggr] \nonumber\\
&=&k( \tilde{X},X^\prime)-R(F_X)\nonumber\\
&=&k(\tilde{X}, X^\prime) - \mb{E}_X[k\tilde{X}, X)], \qquad \forall k(\cdot, X) \in \mc{H}_X. \nonumber
\end{eqnarray}
Which is estimated with the data points $X_1, X_2, \cdots, X_n \in \mc{X}$ as
\begin{eqnarray}
 k(\tilde{X}, X^\prime) - \frac{1}{n}\sum_{i=1}^n k(\tilde{X}, X_i), \qquad \forall k(\cdot, X_i)\in \mc{H}_X,\,\tilde{X},\, X^\prime \in\mc{X}. \nonumber
\end{eqnarray}
\end{Exm}

\begin{Exm}[Kernel cross-raw moment]
 Let  $R(F_{XY}) = \mb{E}_{XY}[ \langle k_X(\cdot, X), f_X \rangle_{\mc{H}_X} \langle k_Y(\cdot, Y), f_Y \rangle_{\mc{H}_Y}] = \mb{E}_{XY}[f_X(X)f_Y(Y)] = \int f_X(X) f_Y(Y)dF_{XY}$. The value of parameter at at $Z^\prime = (X^\prime,       Y^\prime)$  the contamination data, for every point $\tilde{Z} = (\tilde{X},\tilde{Y})$ $W_{XY}^\epsilon= (1 - \epsilon)F_{XY} + \epsilon\Delta_{Z^\prime}$ is given by

\begin{eqnarray}
R[W_{XY}^\epsilon] = R[(1-\epsilon)F_{XY} + \epsilon\Delta_{Z^\prime}]&=& \int f_X(\tilde{X}) f_Y(\tilde{Y})d[(1 - \epsilon)F_{XY} + \epsilon \Delta_{Z^\prime}] \nonumber \\
&=&(1 - \epsilon)\int f_X(\tilde{X}) f_Y(\tilde{Y})dF_{X Y} + \epsilon\int f_X(\tilde{X})f_Y(\tilde{Y})d_{\Delta_{Z^\prime}} \nonumber \\
&=& (1-\epsilon)\int f_X(\tilde{X}) f_Y(\tilde{Y})dF_{XY} +\epsilon f_X(X^\prime)f_Y(Y^\prime)\nonumber\\
&=& (1-\epsilon) R(F_{XY}) + \epsilon f_X(X^\prime)f_Y(Y^\prime) \nonumber
\end{eqnarray}
Thus the IF of $R(F_{XY})$  is given by
\begin{eqnarray}
IF( \cdot, Z^\prime, R, F_{XY}) &=&\lim_{\epsilon\to 0 }\frac{R[W_{XY}^\epsilon] - R(F_{XY})}{\epsilon} \nonumber\\
&=& \lim_{\epsilon\to 0 }\frac{(1-\epsilon)RF_{XY}+\epsilon f_X(X^\prime)f_Y(Y^\prime) - R(F_{XY})}{\epsilon} \nonumber\\
&=& \lim_{\epsilon\to 0 }\biggl[f_X(X^\prime)f_Y(Y^\prime) - R(F_{XY}) \biggr] \nonumber\\
&=& f_X(X^\prime)f_Y(Y^\prime) - R(F_{XY})\nonumber\\
&=& k_X(\tilde{X}, X^\prime) k_Y(\tilde{Y}, Y^\prime) -\mb{E}_{XY}[\langle k_X(\tilde{X}, X), f_X \rangle_{\mc{H}_X} \langle k_Y(\tilde{Y},Y), f_Y \rangle_{\mc{H}_Y}]. \nonumber
\end{eqnarray}
Which is estimated as
\begin{eqnarray}
k_X(X_i, X^\prime) k_Y(X_i, Y^\prime) - \frac{1}{n}\sum_{b=1}^n k_X(X_i,  X_b)k_Y(Y_i, Y_b)\nonumber
\end{eqnarray} 
\end{Exm}

\begin{Exm}[Kernel cross-covariance operator] 
 An cross-covariance operator of $(X, Y)$,  $\Sigma_{YX}: \mc{H}_X \to \mc{H}_Y$  is defined as
\begin{eqnarray}
R(F_{YX}) &=& \langle f_Y, \Sigma_{YX}f_X \rangle_{\mc{H}_Y} =\mb{E}_{XY}[ \langle k_X(\cdot, X)-\mc{M}[F_X], f_X \rangle_{\mc{H}_X} \langle k_Y(\cdot,Y)-\mc{M}[F_Y], f_Y \rangle_{\mc{H}_Y}] \nonumber \\
&=& \mb{E}_{XY}[(f_X(X)- \mb{E}_{X}[f(X)]) (f_Y(Y)- \mb{E}_{Y}[f_Y(Y)])]  \nonumber \\
&=&\mb{E}_{XY}[f_X(X) g_Y(Y)] - \mb{E}_{X}[f_X(X)] \mb{E}_Y[g_Y(Y)] \nonumber
\end{eqnarray}
for $f_X\in \mc{H}_X$ and $f_Y\in \mc{H}_Y$. 
The IF of $R(F_{XY})$ at $Z^{\prime}= (X^\prime, Y^\prime)$  using the rule of IF of complicated statistics  is given by
\begin{eqnarray}
\rm{IF}(\cdot, Z^\prime, R, F_{XY}) &=&  f_X(X^\prime)f_Y(Y^\prime)- \mb{E}_{XY}[f_X(X)f_Y(Y)]- \mb{E}_{Y}[ f_Y(Y)][f(X) -  \mb{E}_{X}[ f_X(X)]]\nonumber \\
&-& \mb{E}_{X}[ f_X(X)][ f_Y(X )-  \mb{E}_{X}[ f_Y(Y)] \nonumber \\
&=& [ f_Y(X)- \mb{E}_{X}[ f_X(X)]][ f_Y(Y)- \mb{E}_{Y}[ f_Y(Y)]]- R(F_{XY}) \nonumber \\
&=&  \langle k_X(\cdot, X)-\mc{M}[F_X],  f_X \rangle_{\mc{H}_X} \langle k_Y(\cdot,Y)\mc{M}[F_Y],  f_Y \rangle_{\mc{H}_Y} \nonumber \\
&-&\mb{E}_{XY}[ \langle k_X(\cdot, X)-\mc{M}[F_X], f_X \rangle_{\mc{H}_X} \langle k_Y(\cdot,Y)-\mc{M}[F_Y],  f_Y \rangle_{\mc{H}_Y}] \nonumber
 \end{eqnarray}
\end{Exm}
Which is estimated with the data points $(X_1, Y_1), (X_2, Y_2), \cdots, (X_n, Y_n) \in \mc{X}\times \mc{Y}$ as
\begin{eqnarray}
 \left[k_X(X_i, X^\prime) - \frac{1}{n}\sum_{b=1}^n k_X(X_i, X_b)\right] \left[k_Y(Y_i, Y^\prime)-\frac{1}{n}\sum_{b=1}^n k_Y(Y_i, Y_b)\right]- \nonumber\\
\frac{1}{n}\sum_{j=1}^n\left[k_X(X_i, X_j) - \frac{1}{n}\sum_{b=1}^n k_X(X_i, X_b)] [k_Y(Y_i, Y_j) - \frac{1}{n}\sum_{d=1}^n k_Y(Y_i, Y_d)\right] \nonumber
\end{eqnarray}

\subsection{Robust kernel cross-covariance operato}

\begin{lem}
\label{lemma1}
Under the assumptions (i) and (ii) the G\^{a}teaux differential of the objective function $J$ at $g_1\in \otimes^r \mc{H}$ and incremental $g_2 \in \otimes^r \mc{H}$  is 
\[\delta J(g_1, g_2)=-\langle G(g_1), g_2\rangle_{\otimes^r \mc{H}},\]
where $G: \otimes^r \mc{H} \to \otimes^r \mc{H}$ is defined as
\[ G(g_1)=\frac{1}{n}\sum_{i=1}^n\phi (\|\otimes^r\vc{\Phi}_c(X_i) - g_1\|_{\otimes^r \mc{H}})\cdot (\otimes^r\vc{\Phi}_c(X_i) - g_1). \]
The necessary condition for $g_1=\hat{f} ^{(r)}$, the kernel central moment element  is  $G(g_1)=\vc{0}$
 \end{lem} 

\begin{lem} 
Under the same assumption of Lemma \ref{lemma1}, $r$th  robust kernel  central moment element (robust kernel CME)  at $X$ is given as
\[\hat{f}_c^{(r)} (X, X, \cdots, X) ^{(h)} = \sum_{i=1}^n w_i^{(h-1)}\tilde{k}(X, X_i) \tilde{k}(X, X_i)\cdots \tilde{k}(X, X_i)
\]
where $w_i^{(h)}=\frac{\varphi(\| \otimes_{b=1}^r\vc{\Phi}_c^{(b)}(X_i)- g_c\|_{\otimes\mc{H}_X})}{\sum_{a=1}^n\varphi(\| \otimes_{b=1}^r\vc{\Phi}_c^{(b)}(X_i)- g_c\|)_{\otimes\mc{H}_X}}\,, \rm{and} \, \varphi(x)=\frac{\zeta^\prime(x)}{x}.$ Putting the different value of $r$, we get the  different robust kernel moment estimates.
 \end{lem} 

\begin{cor}
Under the same assumption of Lemma \ref{lemma1}, kernel CO at $(X,X)$ and kernel CCO at $(X, Y)$ are estimated as  
\[\widehat{\Sigma}_{X X}^{(h)}= \sum_{i=1}^n w_i^{(h-1)}\tilde{k}(X, X_i)\tilde{k}(X, X_i),\qquad
\widehat{\Sigma}_{X Y}^{(h)}= \sum_{b=1}^n w_b^{(h-1)}\tilde{k}(X, X_b) \tilde{k}(Y, Y_i),\]
respectively and  $w_b^{(h)}$ is the same as in  Lemma Eq. (\ref{lemma1}) with $r=2$.
\end{cor}

\subsubsection{Proof of  Theorem $4.1$: Influence function of kernel CCA}
As in \cite{Fukumizu-SCKCCA}, using the cross-covariance operator of (X,Y), $\Sigma_{XY}: \mc{H}_Y\to \mc{H}_X$  we can reformulate the optimization problem of classical kernel canonical correlation (classical kernel CCA)  as follows:
\begin{eqnarray}
\label{ckcca2}
\sup_{\substack{f_{X}\in \mc{H}_X, f_{Y}\in \mc{H}_Y \\ f_{X}\ne 0,\,f_{Y}\ne 0}}\langle f_X,\Sigma_{XY}f_Y\rangle_{\mc{H}_X}\qquad
\text{subject to}\qquad
\begin{cases}
	\langle f_X, \Sigma_{XX}f_X\rangle_{\mc{H}_X}=1,
\\
\langle f_Y, \Sigma_{YY}f_Y\rangle_{\mc{H}_Y}=1.
\end{cases}
\end{eqnarray}
 Using generalized eigenvalue problem, we can derive the solution of Eq. (\ref{ckcca2}) as with liner CCA \citep{Anderson-03}.
\begin{eqnarray}
\label{ckcca3}
\begin{cases}
\Sigma_{XY}f_X - \rho\Sigma_{YY} f_Y = 0,
\\
\Sigma_{XY}f_Y - \rho\Sigma_{XX}f_X = 0. 
\end{cases} \nonumber
\end{eqnarray}
After some simple calculation, we can reset the solution as a single matrix equation  for $f_X$ or $f_Y$.
\begin{eqnarray}
\label{ckcca4}
\begin{cases}
(\Sigma_{XY} \Sigma_{YY}^{-1}\Sigma_{XY} - \rho^2 \Sigma_{XX})f_X = 0,
\\
(\Sigma_{XY} \Sigma_{XX}^{-1}\Sigma_{XY}-\rho^2 \Sigma_{YY})f_Y = 0.
\end{cases}
\end{eqnarray}
The generalized eigenvalue problem  in  Eq. (\ref{ckcca4}) (for simplicity we use first equation only) can be formulated as a simple eigenvalue problem using {\it j}th eigenfunction.
\begin{eqnarray}
\label{ckcca5}
(\Sigma_{XX}^{- \frac{1}{2}} \Sigma_{XY} \Sigma_{YY}^{-1}\Sigma_{YX}\Sigma_{XX}^{- \frac{1}{2}} - \rho_j^2I) \Sigma_{XX}^{\frac{1}{2}}f_{jX}&=&0 \nonumber \\
\Rightarrow (\Sigma_{XX}^{-\frac{1}{2}} \Sigma_{XY} \Sigma_{YY}^{-1}\Sigma_{YX}\Sigma_{XX}^{- \frac{1}{2}} - \rho_j^2I) f_{jX}& = &0
\end{eqnarray}

To use the results of IF of  liner principle components analysis \citep{Tanaka-88},  IF of liner canonical correlation analysis \citep{Romanazii-92}  and  IF of  kernel principle component analysis \citep {Huang-KPCA}  for the finite dimension  and  for the infinite dimension, respectively,  we  convert generalized eigenvalue problem of kernel canonical correlation analysis   into a simple eigenvalue problem. Thus,  we need to find, the IF of  $ \Sigma_{XX}^{- \frac{1}{2}} \Sigma_{XY} \Sigma_{YY}^{-1}\Sigma_{YX}\Sigma_{XX}^{- \frac{1}{2}}$ and henceforth  IF of $\Sigma_{YY}^{-1},  \Sigma_{XX}^{\frac{1}{2}}$ and $\Sigma_{XY}$. Let $\Sigma_{XY}$ be the covariance of the random vectors $k_X(\cdot, X)$ and $k_Y(\cdot, Y)$ on RKHS i.e.,  kernel covariance operator, $\Sigma_{YX}: \mc{H}_X\to\mc{H}_Y$, for all $f_X\in \mc{H}$ and  $f_Y\in \mc{H}_Y$ we have
\begin{align}
 &\mb{E}_{XY}[ \langle f_X, k_X(\cdot, X)-\mc{M}_X\rangle_{\mc{H}_X} \langle  k_Y(\cdot, Y)-\mc{M}_Y, f_Y\rangle_{\mc{H}_Y}]\nonumber\\
 &= \mb{E}_{XY}[ \langle f_Y, ((k_X(\cdot, X)-\mc{M}_X)\otimes(k_Y(\cdot, Y)-\mc{M}_Y)) f_X\rangle_{\mc{H}_Y}]\nonumber\\
&=  \langle f_Y, \mb{E}_{XY}(k_X(\cdot, X) - \mc{M}_X)\otimes(k_Y(\cdot, Y) - \mc{M}_Y)) f_X\rangle_{\mc{H}_Y}\nonumber\\
&=\langle f_Y, \Sigma_{YX}f_X\rangle_{\mc{H}_Y} 
\end{align}
where $\mc{M}_X$ is kernel mean elements in $\mc{H}_X$ and  $\Sigma_{YX}= (k_X(\cdot, X) - \mc{M}_X)\otimes(k_Y(\cdot, Y) - \mc{M}_Y)$, 
since $((T_1\otimes T_2)(x) = \langle x,T_2\rangle T_1$). Using simple algebra we  have at $Z^\prime= (X^\prime, Y^\prime)$

\begin{align} 
&\rm{IF}(\cdot, X^\prime, \Sigma_{XX})= (k_X(\cdot, X^\prime)-\mc{M}_X)  \otimes(k_X(\cdot, X^\prime)-\mc{M}_X)- \Sigma_{XX}, \nonumber\\
&\rm{IF}(\cdot, Y^\prime, \Sigma_{YY})= (k_Y(\cdot, Y^\prime)-\mc{M}_Y) \otimes  (k_Y(\cdot, Y^\prime)-\mc{M}_Y)- \Sigma_{YY},\nonumber\\
&\rm{IF}(\cdot, Z^\prime, \Sigma_{XY})= (k_X(\cdot, X^\prime) -\mc{M}_X)\otimes  (k_Y(\cdot, Y^\prime)-\mc{M}_Y)- \Sigma_{XY} \, \rm{and} \nonumber\\
&\rm{IF}(Z^\prime, X^\prime, 
\Sigma_{XX}^{-\frac{1}{2}} )= \frac{1}{2} [\Sigma_{XX}^{-\frac{1}{2}}- \Sigma_{XX}^{-\frac{1}{2}}(k_X(\cdot, X^\prime)-\mc{M}_X)\otimes  (k_X(\cdot, X^\prime)-\mc{M}_X) \Sigma_{XX}^{-\frac{1}{2}}]. \nonumber
\label{IFIVCOV}
\end{align} 

For simplicity, let us define $\tilde{k}_X (\cdot, X^\prime):= k_X(\cdot, X^\prime)-\mc{M}_X, \, \tilde{k}_Y (\cdot, \vc{y}^\prime):= k_X(\cdot,  Y^\prime) - \mc{M}_Y\,  \rm{and}\, \mb{A}:= \Sigma_{XY} \Sigma_{YY}^{-1}\Sigma_{YX}, \, \mb{B}:= \Sigma_{XX}^{- \frac{1}{2}}\mb{A}\Sigma_{XX}^{- \frac{1}{2}}$, and $\mb{L}= \Sigma_{XX}^{- \frac{1}{2}}(\Sigma_{XX}^{- \frac{1}{2}} \Sigma_{XY} \Sigma_{YY}^{-1} \Sigma_{YX}\Sigma_{XX}^{- \frac{1}{2}}-\rho^2\vc{I})^{-1}\Sigma_{XX}^{- \frac{1}{2}}$ 
Now, 
\begin{align} 
&\rm{IF} (\cdot, Z^\prime, \mb{A})= \rm{IF}(\vc{x}^\prime, Y^\prime, \Sigma_{XY}) \Sigma^{-1}_{YY}\Sigma_{YX} + \Sigma_{XY}  \rm{IF}(X^\prime, Y^\prime, \Sigma_{YY}^{-1})\Sigma_{YX}+ \Sigma_{XY} \Sigma_{YY}^{-1}\rm{IF}(X^\prime, Y^\prime, \Sigma_{XY}) \nonumber\\
&= \left[\tilde{k}_X (\cdot, X^\prime)\otimes \tilde{k}_Y (\cdot, Y^\prime)- \Sigma_{XY}\right]\Sigma_{YY}^{-1}\Sigma_{YX}
+ \Sigma_{XY}\left[\Sigma_{YY}^{-1}-\Sigma_{YY}^{-1}\tilde{k}_Y (\cdot, Y^\prime)\otimes \tilde{k}_Y (\cdot, Y^\prime) \Sigma_{YY}^{-1}\right] \Sigma_{YX}\nonumber\\
&\qquad \qquad\qquad \qquad\qquad \qquad\qquad \qquad+ \Sigma_{XY}\Sigma_{YY}^{-1}\left[\tilde{k}_X (\cdot, X^\prime)\otimes \tilde{k}_Y (\cdot, Y^\prime)- \Sigma_{YX}\right]\nonumber\\
&=2 \Sigma_{XY}\Sigma^{-1}_{YY}\left[\tilde{k}_X(\cdot, X^\prime)\otimes  \bar{k}_Y(\cdot, Y^\prime)- \Sigma_{XY}\right]
+ \Sigma_{XY}\left[\Sigma_{YY}^{-1}-\Sigma_{YY}^{-1}\tilde{k}_Y (\cdot, Y^\prime)\otimes \tilde{k}_Y (\cdot, Y^\prime) \Sigma_{YY}^{-1}\right] \Sigma_{YX}  \nonumber
\end{align} 
Then,
\begin{multline}
\Sigma_{XX}^{-\frac{1}{2}} \rm{IF}(Z^\prime, \mb{A}) \Sigma_{XX}^{-\frac{1}{2}}= 2  \Sigma_{XX}^{-\frac{1}{2}}\Sigma_{XY}\Sigma^{-1}_{YY}[ \tilde{k}_X (\cdot, X^\prime)\otimes \tilde{k}_Y (\cdot, Y^\prime)- \Sigma_{XY}] \Sigma_{XX}^{-\frac{1}{2}}\nonumber\\
+\Sigma_{XX}^{-\frac{1}{2}}\Sigma_{XY}[\Sigma_{YY}^{-1}-\Sigma_{YY}^{-1}[\tilde{k}_Y (\cdot,\vc{y}^\prime)\otimes \tilde{k}_Y (\cdot,\vc{y}^\prime)] \Sigma_{YY}^{-1}] \Sigma_{YX}\Sigma_{XX}^{-\frac{1}{2}} \nonumber
\end{multline}
and
\begin{multline}
 \rm{IF}(X^\prime, \Sigma_{XX}^{-\frac{1}{2}})\mb{A} \Sigma_{XX}^{-\frac{1}{2}}+ \Sigma_{XX}^{-\frac{1}{2}}\mb{A} \rm{IF}(X^\prime, \Sigma_{XX}^{-\frac{1}{2}})
= 2 \rm{IF}(X^\prime, \Sigma_{XX}^{-\frac{1}{2}})\mb{A} \Sigma_{XX}^{-\frac{1}{2}}
= [\Sigma_{XX}^{-\frac{1}{2}}- \Sigma_{XX}^{-\frac{1}{2}}\tilde{k}_X (\cdot, X^\prime)\otimes \tilde{k}_X (\cdot, X^\prime) \Sigma_{XX}^{-\frac{1}{2}}] \mb{A} \Sigma_{XX}^{-\frac{1}{2}}\nonumber
\end{multline}
The  influence of $\mb{B}$ is given by  
\begin{align} 
&\rm{IF} (X^\prime, Y^\prime,\mb{B})= 2\rm{IF} (X^\prime, Y^\prime, \Sigma_{XX}^{-\frac{1}{2}}) \Sigma_{XY} \Sigma_{YY}^{-1}\Sigma_{XY} \Sigma_{XX}^{-\frac{1}{2}}+ \Sigma_{XX}^{-\frac{1}{2}} \rm{IF}(X^\prime, Y^\prime, \mb{A}) \Sigma_{XX}^{-\frac{1}{2}} \nonumber\\
&=  [\Sigma_{XX}^{-\frac{1}{2}}- \Sigma_{XX}^{-\frac{1}{2}}\tilde{k}_X (\cdot, X^\prime)\otimes \tilde{k}_X (\cdot, X^\prime)\Sigma_{XX}^{-\frac{1}{2}}] \Sigma_{XY} \Sigma_{YY}^{-1}\Sigma_{XY}\Sigma_{XX}^{-\frac{1}{2}} \nonumber\\
& \qquad \qquad \qquad\qquad + 2  \Sigma_{XX}^{-\frac{1}{2}}\Sigma_{XY}\Sigma^{-1}_{YY}[ \tilde{k}_X (\cdot, X^\prime)\otimes \tilde{k}_Y (\cdot, Y^\prime) - \Sigma_{XY}] \Sigma_{XX}^{-\frac{1}{2}}\nonumber\\
&\qquad\qquad\qquad\qquad+  \Sigma_{XX}^{-\frac{1}{2}}\Sigma_{XY}[\Sigma_{YY}^{-1} - \Sigma_{YY}^{-1}\tilde{k}_Y (\cdot, Y^\prime)\otimes \tilde{k}_Y (\cdot, Y^\prime) \Sigma_{YY}^{-1}] \Sigma_{YX}\Sigma_{XX}^{-\frac{1}{2}}\nonumber \\
&= - \Sigma_{XX}^{-\frac{1}{2}}\tilde{k}_X (\cdot, X^\prime)\otimes \tilde{k}_X (\cdot, X^\prime) \Sigma_{XX}^{-\frac{1}{2}} \Sigma_{XY} \Sigma_{YY}^{-1}\Sigma_{XY}\Sigma_{XX}^{-\frac{1}{2}} \nonumber \\
& \qquad \qquad \qquad\qquad+ 2  \Sigma_{XX}^{-\frac{1}{2}}\Sigma_{XY}\Sigma^{-1}_{YY} \tilde{k}_X (\cdot, X^\prime)\otimes \tilde{k}_Y (\cdot, X^\prime) \Sigma_{XX}^{-\frac{1}{2}}\nonumber \\
&\qquad\qquad \qquad\qquad-  \Sigma_{XX}^{-\frac{1}{2}}\Sigma_{XY}\Sigma_{YY}^{-1}\tilde{k}_Y (\cdot, Y^\prime)\otimes \tilde{k}_Y (\cdot, Y^\prime) \Sigma_{YY}^{-1} \Sigma_{YX}\Sigma_{XX}^{-\frac{1}{2}}
\end{align}
We convert generalized eigenvalue problem as a eigenvalue problem and use the Lemma 1 of \cite{Huang-KPCA} to define the IF of kernel CC, $\rho_j^2$ and  kernel CVs, $f_X(X)$ and, $f_Y(Y)$. Then the  IF of kernel $\rho_j^2$ is defined as
\begin{align} 
&\rm{IF} (Z^\prime,\rho_j^2)= \langle\tilde{f}_{jX}, \rm{IF} (Z^\prime, \mb{B}) \tilde{f}_{jX}\rangle_{\mc{H}_X\otimes \mc{H}_Y}\nonumber\\
&=-\langle \tilde{f}_{jX}, \Sigma_{XX}^{-1}\tilde{k}_X (\cdot, X^\prime)\otimes \tilde{k}_X (\cdot, X^\prime) \Sigma_{XX}^{-\frac{1}{2}} \Sigma_{XY} \Sigma_{YY}^{-1}\Sigma_{XY}\Sigma_{XX}^{-\frac{1}{2}} \tilde{f}_{jX}\rangle_{\mc{H}_X\otimes \mc{H}_X}\nonumber\\
& \qquad \qquad \qquad\qquad+\ 2\langle  \tilde{f}_{jX}, \Sigma_{XX}^{-\frac{1}{2}}\Sigma_{XY}\Sigma^{-1}_{YY}\tilde{k}_X (\cdot, X^\prime)\otimes \tilde{k}_Y (\cdot, Y^\prime) \Sigma_{XX}^{-\frac{1}{2}} \tilde{f}_{jX}\rangle_{\mc{H}_X\otimes \mc{H}_Y}\nonumber\\
&\qquad \qquad \qquad\qquad-\langle \tilde{f}_{jX}^T,  \Sigma_{XX}^{-\frac{1}{2}}\Sigma_{XY}\Sigma_{YY}\tilde{k}_Y (\cdot, Y^\prime)\otimes \tilde{k}_Y (\cdot, Y^\prime)\Sigma_{YX}\Sigma_{XX}^{-\frac{1}{2}} \tilde{f}_{jX}\rangle_{\mc{H}_Y\otimes \mc{H}_Y}\nonumber\\
&=-\langle\tilde{f}_{jX}, \Sigma_{XX}^{-1}\tilde{k}_X (\cdot, X^\prime)\otimes \tilde{k}_X (\cdot, X^\prime) \Sigma_{XX}^{-\frac{1}{2}} \Sigma_{XY} \Sigma_{YY}^{-1}\Sigma_{XY}\Sigma_{XX}^{-\frac{1}{2}} \tilde{f}_{jX}\rangle_{\mc{H}_X\otimes \mc{H}_X}\nonumber\\
& \qquad \qquad \qquad\qquad +2 \langle  \tilde{f}_{jX}^T, \Sigma_{XX}^{-\frac{1}{2}}\Sigma_{XY}\Sigma^{-1}_{YY}\tilde{k}_X (\cdot, X^\prime)\otimes \tilde{k}_Y (\cdot, Y^\prime) \Sigma_{XX}^{-\frac{1}{2}} \tilde{f}_{jX}\rangle_{\mc{H}_X\otimes \mc{H}_Y}\nonumber\\
& \qquad \qquad \qquad\qquad -\langle\tilde{f}_{jX}, \Sigma_{XX}^{-\frac{1}{2}}\Sigma_{XY}[\Sigma_{YY}\tilde{k}_Y (\cdot, Y^\prime)\otimes \tilde{k}_Y (\cdot, Y^\prime) \Sigma_{YY}^{-1}] \Sigma_{YX}\Sigma_{XX}^{-\frac{1}{2}} \tilde{f}_{jX}\rangle_{\mc{H}_Y\otimes \mc{H}_Y} 
\label{IFKCCA1}
\end{align} 
For simplicity we calculate in parts of Eq. (\ref{IFKCCA1}).  The first part  derive as 
\begin{align} 
&\langle\tilde{f}_{jX}, \Sigma_{XX}^{-1}\tilde{k}_X (\cdot, X^\prime)\otimes \tilde{k}_X (\cdot, X^\prime)  \Sigma_{XX}^{-\frac{1}{2}} \Sigma_{XY} \Sigma_{YY}^{-1}\Sigma_{XY}\Sigma_{XX}^{-\frac{1}{2}} \tilde{f}_{jX}\rangle_{\mc{H}_X\otimes \mc{H}_X} \nonumber\\
&= \langle \Sigma_{XX}^{-\frac{1}{2}}\tilde{f}_{jX},\tilde{k}_X (\cdot, X^\prime)\otimes \tilde{k}_X (\cdot, X^\prime)  \Sigma_{XX}^{-\frac{1}{2}} \Sigma_{XY} \Sigma_{YY}^{-1}\Sigma_{XY}\Sigma_{XX}^{-\frac{1}{2}}\Sigma_{XX}^{-\frac{1}{2}} \tilde{f}_{jX}\rangle_{\mc{H}_X\otimes \mc{H}_X} \nonumber\\
&= \langle f_{jX}, \tilde{k}_X(\cdot, X^\prime)\rangle_{\mc{H}_X} \langle \tilde{k}_X(\cdot, X^\prime), \Sigma_{XX}^{-\frac{1}{2}} \Sigma_{XY} \Sigma_{YY}^{-1}\Sigma_{XY}\Sigma_{XX}^{-\frac{1}{2}} f_{jX}\rangle_{\mc{H}_X}\nonumber\\
&= \rho_j^2 \bar{f}_{jX}^2(X^\prime), 
\label{IFKCCA2}
\end{align} 
in the last equality, we use   Eq. (\ref{ckcca5}). The 2nd part of  the Eq. (\ref{IFKCCA1}) derive as  
\begin{align} 
&\langle f_{jX}, \Sigma_{XX}^{-\frac{1}{2}}\Sigma_{XY}\Sigma^{-1}_{YY}[ \tilde{k}_X (\cdot, X^\prime)\otimes \tilde{k}_Y (\cdot, Y^\prime)] \Sigma_{XX}^{-\frac{1}{2}} f_{jX}\rangle_{\mc{H}_X\otimes \mc{H}_Y} \nonumber\\
&= \langle \Sigma_{XX}^{-\frac{1}{2}} \tilde{f}_{jX}, \tilde{k}_X (\cdot, X^\prime)\otimes \tilde{k}_Y (\cdot, Y^\prime)\Sigma_{XY}\Sigma^{-1}_{YY} \Sigma_{XX}^{-\frac{1}{2}} f_{jX}\rangle_{\mc{H}_X\otimes \mc{H}_Y} \nonumber\\
&=\langle f_{jX},\tilde{k}_X (\cdot, X^\prime)\otimes \tilde{k}_Y (\cdot, Y^\prime) \Sigma_{XY}\Sigma^{-1}_{YY}  f_{jX}\rangle_{\mc{H}_X\otimes \mc{H}_Y} \nonumber\\
&=  \rho_j \langle f_{jX}, \tilde{k}_X (\cdot, X^\prime)\rangle_{\mc{H}_X} \langle \tilde{k}_Y (\cdot, Y^\prime),  f_{jY}\rangle_{\mc{H}_Y}  
 \nonumber\\
&=  \rho_j \bar{f}_{jX}( X^\prime) \bar{f}_{jY}(Y^\prime), 
\label{IFKCCA3}
\end{align} 
in the last second equality, we use   Eq.(\ref{ckcca3}). Similarly, we can write  the 3rd term as 
\begin{align}
\label{IFKCCA4}
\langle \tilde{f}_{jX},  \Sigma_{XX}^{-\frac{1}{2}}\Sigma_{XY}[\Sigma_{YY}\tilde{k}_Y (\cdot, Y^\prime)\otimes \tilde{k}_Y (\cdot, Y^\prime)  \Sigma_{YY}^{-1}] \Sigma_{YX}\Sigma_{XX}^{-\frac{1}{2}} \tilde{f}_{jX}\rangle_{\mc{H}_Y\otimes \mc{H}_Y} =   \rho_j^2 \bar{f}_{jY}^2(Y^\prime)
\end{align} 
where  $\bar{f}_{jX}= f_{jX}(X^\prime)= \langle f_{jX}, \tilde{k}_X (\cdot, X^\prime)$ and similar  for $\bar{f}_{jY}$.
Therefore, substituting Eq. (\ref{IFKCCA2}), (\ref{IFKCCA3})  and (\ref{IFKCCA4}) into Eq. (\ref{IFKCCA1})  the IF  of kernel CC is given by 
\begin{align}
\rm{IF} (X^\prime, Y^\prime,\rho_j)= - \rho_j^2 \bar{f}_{jX}^2(Y^\prime)+2 \rho_j\bar{f}_{jX}(X^\prime) \bar{f}_{jY}(Y^\prime)  -\rho_j^2 \bar{f}_{jY}^2(Y^\prime)
\end{align} 
Now we derive the IF of  kernel Cvs. To this end first we need to derive
\begin{align}
\rm{IF} (X^\prime, f_{jx})= \rm{IF} (X^\prime, \Sigma_{XX}^{-\frac{1}{2}}f_{jX})
= \Sigma_{XX}^{-\frac{1}{2}} \rm{IF} (X^\prime, f_{jX})+\rm{IF} (X^\prime, \Sigma_{XX}^{-\frac{1}{2}})f_{jX}
\label{CV1}
\end{align} 
By the first term of Eq. (\ref{CV1}) we have 
\begin{align}
&\Sigma_{XX}^{-\frac{1}{2}} \rm{IF} (X^\prime,  Y^\prime,f_{jX})= \Sigma_{XX}^{-\frac{1}{2}} ( \mb{B}-\rho^2\vc{I})^{-1}\rm{IF} (X^\prime, Y^\prime,\mb{B})f_{jX} \nonumber\\
&= - \Sigma_{XX}^{-\frac{1}{2}} ( \mb{B}-\rho^2\vc{I})^{-1}\big[ \Sigma_{XX}^{-\frac{1}{2}}\tilde{k}_X(\cdot, X^\prime)\otimes  \tilde{k}_X(\cdot, X^\prime)\Sigma_{XX}^{-\frac{1}{2}} \Sigma_{XY} \Sigma_{YY}^{-1}\Sigma_{XY}\Sigma_{XX}^{-\frac{1}{2}}\nonumber\\
&+ 2  \Sigma_{XX}^{-\frac{1}{2}}\Sigma_{XY}\Sigma^{-1}_{YY}\tilde{k}_X(\cdot, X^\prime)\otimes  \tilde{k}_Y(\cdot, Y^\prime) \Sigma_{XX}^{-\frac{1}{2}}-  \Sigma_{XX}^{-\frac{1}{2}}\Sigma_{XY}\Sigma_{YY}\tilde{k}_Y(\cdot, Y^\prime) \Sigma^{-1}_{YY}\Sigma_{YX}\Sigma_{XX}^{-\frac{1}{2}}\big]\bar{f}_{jX}
\label{CV2}
\end{align}

We derive each terms of Eq. (\ref{CV2}), respectively. The first term of  
Eq. (\ref{CV2}) is given by
\begin{align}
&\Sigma_{XX}^{-\frac{1}{2}} (\mb{B}-\rho^2\vc{I})^{-1}[ \Sigma_{XX}^{-\frac{1}{2}} \langle \tilde{k}_X(\cdot, X^\prime)\otimes \tilde{ k}_X(\cdot, X^\prime),  \Sigma_{XX}^{-\frac{1}{2}} \Sigma_{XY} \Sigma_{YY}^{-1}\Sigma_{XY}\Sigma_{XX}^{-\frac{1}{2}} f_{jX}\rangle \nonumber\\
&=\mb{L}\langle\tilde{ k}_X(\cdot, X^\prime)\otimes \tilde{ k}_X(\cdot, X^\prime),  \Sigma_{XX}^{-\frac{1}{2}} \Sigma_{XY} \Sigma_{YY}^{-1}\Sigma_{YX} f_{jX}\rangle \nonumber\\
&= \mb{L} \rho_j^2\langle\tilde{ k}_X(\cdot, X^\prime)\otimes \tilde{ k}_X(\cdot, \vc{x}^\prime),  f_{jX}\rangle  \\ \nonumber
&= \mb{L} \rho_j^2\langle\tilde{ k}_X(\cdot, X^\prime)\otimes \tilde{ k}_X(\cdot, X^\prime),  f_{jX}\rangle  \\ \nonumber
&= \mb{L} \rho_j^2\bar{f}(X^\prime) \tilde{k}(\cdot, X^\prime) \nonumber
\end{align}
2nd term of Eq. (\ref{CV2}) is 
\begin{align}
&2\mb{L}  \Sigma_{XX}^{-\frac{1}{2}}\Sigma_{XY}\Sigma^{-1}_{YY}\tilde{k}_X(\cdot, X^\prime) \tilde{k}_Y(\cdot, Y^\prime) \Sigma_{XX}^{-\frac{1}{2}} f_{jX} \nonumber\\
&= \mb{L} \langle  \Sigma_{XX}^{-\frac{1}{2}}\Sigma_{XY}\Sigma^{-1}_{YY} f_{jX},  \tilde{k}_Y(\cdot, Y^\prime)\rangle \tilde{k}_X(\cdot, X^\prime) + \mb{L} \langle \Sigma_{XX}^{-\frac{1}{2}}\Sigma_{XY}\Sigma^{-1}_{YY}\tilde{k}_X(\cdot, X^\prime),  f_{jX}\rangle \tilde{k}_Y(\cdot, Y^\prime)\nonumber\\   
&= \mb{L}  \rho_j \bar{f}_{jY}(Y^\prime)(\tilde{k}_X(\cdot, X^\prime)+ \mb{L} \Sigma_{XX}^{-\frac{1}{2}}\Sigma_{XY}\Sigma^{-1}_{YY} \bar{f}_{jX} (X^\prime) \tilde{k}_Y(\cdot, Y^\prime)   \nonumber
\end{align}
and the 3rd term of   Eq. (\ref{CV2}) is 
\begin{eqnarray}
\mb{L} \Sigma_{XY}\Sigma^{-1}_{YY}\tilde{k}_Y(\cdot, Y^\prime) \otimes \tilde{k}_Y(\cdot, Y^\prime) \Sigma_{YY}^{-1} \Sigma_{YX}\Sigma_{XX}^{-\frac{1}{2}}]f_{jX} 
&=&\mb{L} \langle  \Sigma_{XY}\Sigma^{-1}_{YY}\tilde{k}_Y(\cdot, Y^\prime) ,  \Sigma_{YY}^{-1}\Sigma_{YX}f_{jX}\rangle \tilde{k}_Y(\cdot, Y^\prime)\rangle \nonumber\\
&=&\mb{L} \langle  \Sigma_{XY}\Sigma^{-1}_{YY}\tilde{k}_Y(\cdot, Y^\prime) , \rho _jf_{jX}\rangle \tilde{k}_Y(\cdot, Y^\prime) \rangle\nonumber\\
&=&\mb{L} \rho _j   \Sigma_{XY}\Sigma^{-1}_{YY} \bar{f}_{jY} (Y^\prime) \tilde{k}_Y(\cdot, Y^\prime) \nonumber 
\end{eqnarray}

By substituting the  above three equations into Eq. (\ref{CV2})  we have
\begin{eqnarray}
 \Sigma_{XX}^{-\frac{1}{2}} \rm{IF} (\cdot, Z^\prime, f_{jX})&=& \Sigma_{XX}^{-\frac{1}{2}} ( \mb{B} - \rho^2\vc{I})^{-1}\rm{IF} (\cdot, Z^\prime,\mb{B})f_{jX} \nonumber\\
&=&-\rho_j (\bar{f}_{jY}(Y^\prime)-\rho_j \bar{f}_{jX}(X^\prime))\mb{L} \tilde{k} (\cdot, X^\prime) -  (\bar{f}_{jX}(X^\prime) - \rho_j \bar{f}_{jY}(Y^\prime)) \Sigma_{XY}\Sigma^{-1}_{YY} \tilde{k}_Y(\cdot, Y^\prime)\nonumber\\
\label{CV6}
\end{eqnarray}
The 2nd term of the Eq. (\ref{CV1}) is give by
\begin{align}
&\rm{IF} (X^\prime, \Sigma_{XX}^{-\frac{1}{2}}) f_{jX} \nonumber \\
&= - \langle f_{jX}, \Sigma_{XX}^{-1} f_{jX}\rangle  \Sigma_{XX}^{-\frac{1}{2}} \rm{IF} (X^\prime, \Sigma_{XX}^{\frac{1}{2}}) \Sigma_{XX}^{-\frac{1}{2}} \tilde{f}_{jX} \nonumber \\
&=  \langle f_{jX}, \Sigma_{XX}^{-\frac{1}{2}} \rm{IF} (X^\prime, \Sigma_{XX}^{\frac{1}{2}}) f_{jX}\rangle f_{jX}\nonumber \\
&= -\frac{1}{2}[\langle  f_{jX}, \Sigma_{XX}^{-\frac{1}{2}} \rm{IF} (X, \Sigma_{XX}^{\frac{1}{2}}) f_{jX}\rangle + \langle  f_{jX},  \rm{IF} (X^\prime, \Sigma_{XX}^{\frac{1}{2}}) \Sigma_{XX}^{-\frac{1}{2}} f_{jX}\rangle]f_{jX} \nonumber \\
&=  - \frac{1}{2}[\langle  f_{jX}, \Sigma_{XX}^{-\frac{1}{2}} \rm{IF} (X^\prime, \Sigma_{XX}) f_{jX}\rangle]f_{jX} \nonumber \\
&= -\frac{1}{2}[ \langle f_{jX}, (\tilde{k}_X(\cdot, X^\prime)-\Sigma_{XX} f_{jX})]f_{jX} \nonumber \\
&= -\frac{1}{2}[\bar{f}_{jX}(X^\prime)- \langle f_{jX},\Sigma f_{jX}\rangle]f_{jX}\nonumber\\
&= \frac{1}{2}[1- \bar{f}_{jX}(X^\prime)] f_{jX}
\label{CV7}
\end{align}
Therefore, substituting  Eq. (\ref{CV6})  and  Eq. (\ref{CV7})   into  Eq. (\ref{CV1}) we get the IF  of kernel canonical variate (CV) of 
\begin{multline}
 \rm{IF} (\cdots, X^\prime, Y^\prime,f_{jX})=-\rho_j (\bar{f}_{jY}(Y^\prime)-\rho_j \bar{f}_{jX}(X^\prime))\mb{L} \tilde{k} (\cdot, X^\prime)-  (\bar{f}_{jX}(X^\prime)-\rho_j \bar{f}_{jY}(Y^\prime))\mb{L} \Sigma_{XY}\Sigma^{-1}_{YY} \tilde{k}_Y(\cdot, Y^\prime)\\ +\frac{1}{2}[1- \bar{f}^2(Y^\prime)]f_{jX} \nonumber
\end{multline}
Similarly, we can derive   $\rm{IF} (\cdot, X^\prime, Y^\prime,f_{jY})$.

Let $(\vc{X}_i, \vc{Y}_i)_{i=1}^n$ be a sample from the distribution $F_{XY}$.  The empirical estimator of Eq. (\ref{ckcca2}) and Eq. (\ref{ckcca4}) are 
\begin{eqnarray}
\sup_{\substack{f_{X}\in \mc{H}_X,f_{Y}\in \mc{H}_X \\ f_{X}\ne 0,\,f_{Y}\ne 0}}\langle f_Y,\hat{\Sigma}_{YX}f_X\rangle_{\mc{H}_Y}\qquad
\text{subject to}\qquad
\begin{cases}
\langle f_X, (\hat{\Sigma}_{XX} + \kappa\vc{I}) f_X\rangle_{\mc{H}_X} = 1,
\\
\langle f_Y, (\hat{\Sigma}_{YY}+\kappa\vc{I})f_Y\rangle_{\mc{H}_Y} = 1,
\end{cases}
\label{ckcca7}
\end{eqnarray}

\begin{eqnarray}
\label{ckcca8}
\begin{cases}
(\hat{\Sigma}_{XY} (\hat{\Sigma}_{YY} + \kappa\vc{I})^{-1}\hat{\Sigma}_{XY} - \rho^2 (\hat{\Sigma}_{XX}+\kappa\vc{I}))f_X = 0,
\\
(\hat{\Sigma}_{YX} (\hat{\Sigma}_{XX} + \kappa\vc{I})^{-1}\hat{\Sigma}_{YX} - \rho^2 (\hat{\Sigma}_{YY}+\kappa\vc{I}))f_Y = 0,
\end{cases}
\end{eqnarray}
respectively. 

Using the above equations, the empirical  IF (EIF) of kernel CC and  kernel CVs  at $Z^\prime = (X^\prime, Y^\prime)$ for all points $Z_i=(X_i, Y_i)$ are $\rm{EIF} (Z^\prime, \rho_j^2)= \hat{\rm{IF}} (Z^\prime, \hat{\rho}_j^2),
\rm{EIF} (\cdot, Z^\prime, f_{jX}) = \hat{\rm{IF}} ( Z_i, Z^\prime, f_{jX}), 
 \rm{EIF} (\cdot, \vc{Z}^\prime, f_{jY})= \hat{\rm{EIF}} (Z_i, Z^\prime, \hat{f}_{jY})$, respectively. 

\end{document}